%% file: MM20_1326.tex
\documentclass[sigconf,screen,balance]{acmart}

\usepackage{hyperref}
\usepackage{gensymb,graphics,subfigure,inputenc,setspace}
\usepackage[linesnumbered,algo2e,boxed]{algorithm2e}
\usepackage[figuresright]{rotating}
\usepackage{textcomp,subfigure,multirow,upgreek}
\usepackage{booktabs}
\usepackage{mdwlist}
\usepackage{amsmath}
\usepackage{amssymb}

\usepackage{ragged2e}
\usepackage{bm}
\usepackage{fontawesome}
\graphicspath{{Figures/}}

\usepackage{caption}
\usepackage{enumitem}
\setitemize{noitemsep,topsep=0pt,parsep=0pt,partopsep=0pt}

\acmSubmissionID{1326}

\copyrightyear{2020}
\acmYear{2020}
\setcopyright{acmcopyright}
\acmConference[MM '20]{Proceedings of the 28th ACM International Conference on Multimedia}{October 12--16, 2020}{Seattle, WA, USA}
\acmBooktitle{Proceedings of the 28th ACM International Conference on Multimedia (MM '20), October 12--16, 2020, Seattle, WA, USA}
\acmPrice{15.00}
\acmDOI{10.1145/3394171.3415280}
\acmISBN{978-1-4503-7988-5/20/10}

\begin{document}
\fancyhead{}
\title{Dual Attention GANs for Semantic Image Synthesis}

\author{Hao Tang$^1$,\, Song Bai$^2$,\, Nicu Sebe$^{13}$}
\affiliation{%
	\institution{$^1$DISI, University of Trento \\
	$^2$Department of Engineering Science, University of Oxford \\
	$^3$Huawei Research Ireland}
{hao.tang@unitn.it,\, songbai.site@gmail.com,\, sebe@disi.unitn.it}
}



\input{0Abstract}




\maketitle

\input{1Introduction}

\input{2RelatedWork}
\input{3Method}
\input{4Experiments}
\input{5Conclusion}

\clearpage
\bibliographystyle{ACM-Reference-Format}
\bibliography{sample-base}

\clearpage

\input{supplementary}

\end{document}

%% file: 0Abstract.tex
\begin{abstract}
In this paper, we focus on the semantic image synthesis task that aims at transferring semantic label maps to photo-realistic images. Existing methods lack effective semantic constraints to preserve the semantic information and ignore the structural correlations in both spatial and channel dimensions, leading to unsatisfactory blurry and artifact-prone results. To address these limitations, we propose a novel Dual Attention GAN (DAGAN) to synthesize photo-realistic and semantically-consistent images with fine details from the input layouts without imposing extra training overhead or modifying the network architectures of existing methods. We also propose two novel modules, i.e., position-wise Spatial Attention Module (SAM) and scale-wise Channel Attention Module (CAM), to capture semantic structure attention in spatial and channel dimensions, respectively. Specifically, SAM selectively correlates the pixels at each position by a spatial attention map, leading to pixels with the same semantic label being related to each other regardless of their spatial distances. Meanwhile, CAM selectively emphasizes the scale-wise features at each channel by a channel attention map, which integrates associated features among all channel maps regardless of their scales. We finally sum the outputs of SAM and CAM to further improve feature representation. Extensive experiments on four challenging datasets show that DAGAN achieves remarkably better results than state-of-the-art methods, while using fewer model parameters. The source code and trained models are available at
\url{https://github.com/Ha0Tang/DAGAN}.
\end{abstract}

%% file: 1Introduction.tex
\section{Introduction}
\label{sec:intro}

\begin{figure} \small
  \includegraphics[width=1\linewidth]{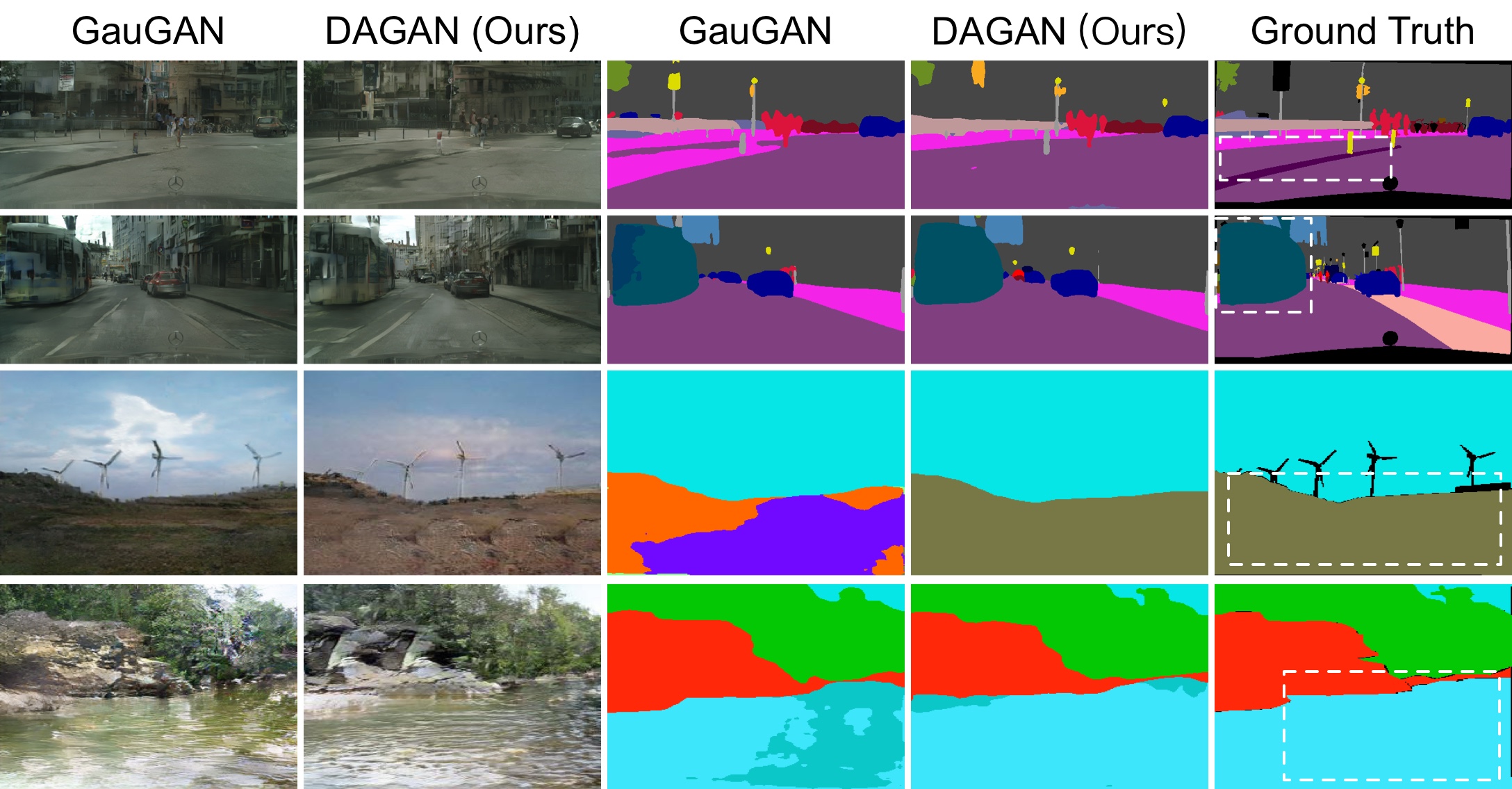}
  \caption{Visualization of generated semantic maps compared with those from GauGAN \cite{park2019semantic} on Cityscapes (\textit{top}) and ADE20K (\textit{bottom}). Equipped with semantic attention modeling in both spatial and channel dimensions, the proposed DAGAN can achieve mutual gains within the regions with the same semantic label regardless of the distances, thus improving intra-class semantic consistency. Most improved regions are highlighted in the ground truths with white dash boxes.}
  \label{fig:seg}
  \vspace{-0.2cm}
\end{figure}

\begin{figure*}[!ht] \small
	\centering
	\includegraphics[width=1\linewidth]{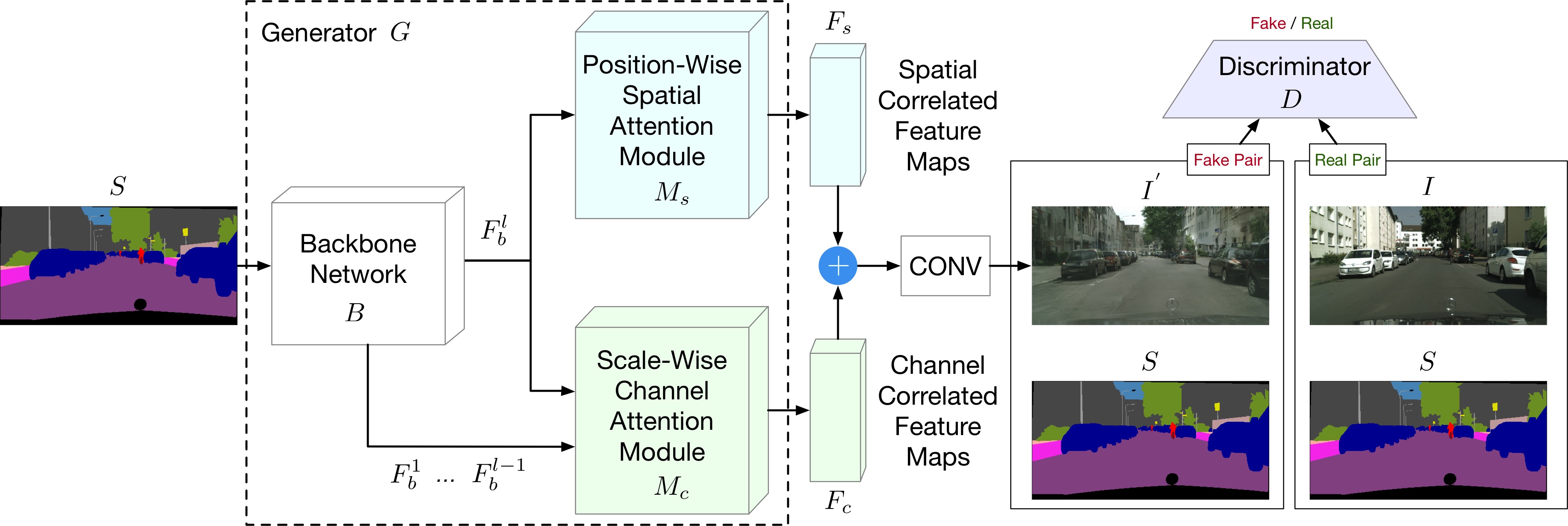}
	\caption{Overview of the proposed DAGAN, which contains a generator $G$ and discriminator $D$. 
	$G$ consists of a backbone network $B$, a position-wise Spatial Attention Module $M_s$ and a scale-wise Channel Attention Module $M_c$. 
	$M_s$ and $M_c$ aim to model the semantic attention in both spatial and channel dimensions for generating semantically-consistent images.
	$D$ aims to distinguish the generated label-image pair from the real one. All of these components are trained in an end-to-end fashion. The symbol $\oplus$ denotes element-wise addition operation.}
	\label{fig:method}
	\vspace{-0.2cm}
\end{figure*}

In this paper, we aim to address the semantic image synthesis task that generates realistic images conditioned on input layouts. 
This has been widely investigated in the recent years \cite{chen2017photographic,isola2017image,qi2018semi,wang2018high,park2019semantic,liu2019learning}.
Most existing methods typically use Generative Adversarial Networks (GANs)~\cite{goodfellow2014generative} to learn the translation mapping from semantic layouts to realistic images. 
For instance, 
Wang et al. \cite{wang2018high} adopt multi-scale generators and discriminators to generate high-resolution images. 
Park et al. \cite{park2019semantic} propose a novel spatially-adaptive normalization for generating realistic images. 
Despite the interesting exploration of these methods, we can still see blurriness and artifacts in the generated images (see Fig.~\ref{fig:seg}), which is mainly due to two reasons.
First, these methods lack effective semantic constraints to maintain the semantic information of the input semantic label.
Second, these methods ignore the semantic correlations between the spatial pixels and channel features which cause intra-class semantic inconsistencies such as the roads, buses, lands and waters generated by GauGAN in Fig.~\ref{fig:seg}.

To solve these limitations, we propose a novel Dual Attention GAN (DAGAN) and two novel modules, i.e., Position-Wise Spatial Attention Module (SAM) and Scale-Wise Channel Attention Module (CAM).
Spatial and channel selections are two crucial factors for translating the input layout to a realistic image. 
Thus both SAM and CAM aim to explore the semantic attention in spatial and channel dimensions for generating high-quality and semantically-consistent images. 
In particular, SAM selectively correlates the pixels at each position by a spatial attention map, leading pixels with the same label to relate to each other regardless of their spatial distances. 
Simultaneously, CAM selectively emphasizes scale-wise features at each channel by a channel attention map integrating associated features among all channel maps regardless of their scales.

Differently from us, the spatial attention module proposed in~\cite{cai2019dualattn} is a non-local model, which requires computing the correlation between every two pixels in the feature map, leading to expensive computational costs and thus limiting its applicability. At the same time, the channel attention module proposed in \cite{cai2019dualattn} did not use multi-scale features, which are quite important for generating small-objects in the semantic image synthesis tasks.

Finally, we sum the outputs of both SAM and CAM to further improve the feature representation.
Notably, both SAM and CAM can be readily applied to existing GAN frameworks without imposing training overheads or modifying network architectures.

\begin{sloppypar}
We perform comprehensive experiments on four challenging datasets with different image resolutions, i.e., ADE20K \cite{zhou2017scene} ($256 {\times} 256$), Cityscapes \cite{cordts2016cityscapes} ($512 {\times} 256$), CelebAMask-HQ \cite{CelebAMask-HQ} ($512 {\times} 512$) and Facades \cite{tylevcek2013spatial} ($1024 {\times} 1024$). 
Both qualitative and quantitative results show that the proposed DAGAN is able to produce remarkably better results than existing models including CRN \cite{chen2017photographic}, SIMS \cite{qi2018semi}, Pix2pixHD \cite{wang2018high}, GauGAN~\cite{park2019semantic} and
CC-FPSE~\cite{liu2019learning}, regarding both the visual fidelity and the alignment with the input layouts.
\end{sloppypar}

Overall, the contributions of our paper are:
\begin{itemize}[leftmargin=*]
	\item We propose a novel Dual Attention GAN (DAGAN) for the challenging task of semantic image synthesis, which can effectively model the semantic attention in both spatial and channel dimensions for improving the ability of feature representations.
	\item We design two novel modules, i.e., position-wise Spatial Attention Module (SAM) and scale-wise Channel Attention Module (CAM), to learn the spatial and channel attention of local features, respectively. Both significantly improve the generation results by modeling intra-class correlations. 
	Moreover, both modules are lightweight and general modules, and can be seamlessly integrated into any existing GAN-based architectures to strengthen the feature representation with negligible overheads. 
	\item We extensively evaluate the proposed DAGAN to confirm that it achieves new state-of-the-art performance on different datasets with different image resolutions, i.e., ADE20K \cite{zhou2017scene} ($256 {\times} 256$), Cityscapes~\cite{cordts2016cityscapes} ($512 {\times} 256$), CelebAMask-HQ \cite{CelebAMask-HQ} ($512 {\times} 512$) and Facades~\cite{tylevcek2013spatial} ($1024 {\times} 1024$), while using significantly fewer model parameters compared with CC-FPSE~\cite{liu2019learning}. Thus it presents new strong baselines for the research community.
\end{itemize}

%% file: 2RelatedWork.tex
\section{Related Work}
\noindent \textbf{Generative Adversarial Networks (GANs)} \cite{goodfellow2014generative} are widely used techniques to learn a complex and  high-dimensional data distribution for generating new images \cite{karras2018progressive,brock2019large,karras2019style,shaham2019singan,zhang2019self}.
\begin{figure*}[!ht]\small
	\centering
	\includegraphics[width=1\linewidth]{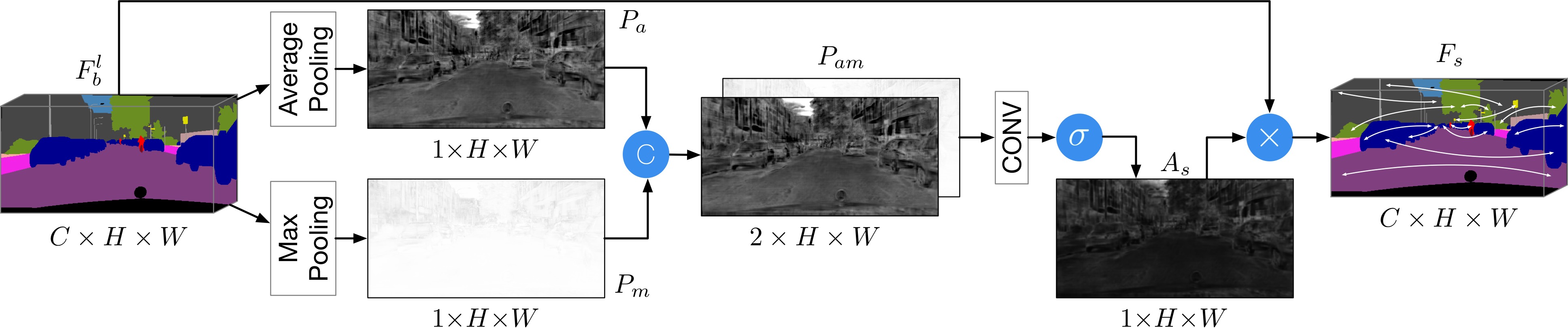}
	\caption{Overview of the proposed position-wise Spatial Attention Module (SAM), which aims to learn the position-wise attention between spatial pixels with the same label regardless of their spatial distances. The symbols $\otimes$, $\textcircled{c}$ and $\textcircled{$\sigma$}$ denote element-wise multiplication, channel-wise concatenation, and Sigmoid($\cdot$) operation, respectively.}
	\label{fig:spatial}
	\vspace{-0.2cm}
\end{figure*}
A vanilla GAN consists of a generator and a discriminator.
The generator aims to produce realistic images to fool the discriminator while the discriminator aims to accurately tell whether an image is real or generated.
Based on GANs, Mirza and Osindero proposed Conditional GANs (CGANs) \cite{mirza2014conditional} by incorporating conditional guidance information to generate user-specific images. 
Conditional guidance information can be category labels \cite{choi2018stargan,zhang2018sparsely,lu2018attribute,wu2019relgan}, text descriptions
\cite{zhang2017stackgan,li2019controllable,yu2019multi}, human pose \cite{balakrishnan2018synthesizing,tang2019cycle,chan2019everybody,neverova2018dense,tang2020xinggan}, segmentation maps \cite{park2019semantic,tang2019multi,wang2018high,liu2020exocentric,wang2019example,gu2019mask,zhang2020dual,tang2020local} and attention maps \cite{mejjati2018unsupervised,kim2020u,tang2019attention,chen2018attention,pumarola2018ganimation}.

\noindent \textbf{Semantic Image Synthesis} aims to turn semantic label maps into photo-realistic images \cite{chen2017photographic,qi2018semi,wang2018high,park2019semantic,liu2019learning}. 
For instance, Park et al.~\cite{park2019semantic} propose a novel spatially-adaptive normalization to preserve semantic information of input labels for generating realistic images. Although GauGAN \cite{park2019semantic} has achieved promising results, we still observe unsatisfactory aspects mainly in the generated scene details and layouts (see Fig.~\ref{fig:seg}), which we believe are mainly due to the problem of missing of spatial and channel semantic information associated with deep network operations.

To tackle this limitation, we propose two novel modules, i.e., position-wise Spatial Attention Module (SAM) and scale-wise Channel Attention Module (CAM), which try to enhance features in both spatial and channel dimensions for generating semantically-consistent results. 
To the best of our knowledge, this idea has not been considered by any existing semantic image generation method.

\noindent \textbf{Semantic Attention Modeling} aims to model
semantic dependencies of distant regions and has widely been applied in many tasks such as semantic segmentation \cite{lin2016efficient,fu2019dual,ding2020lanet}, depth estimation \cite{xu2018structured}, sentiment classification \cite{lin2017structured}, machine
translation \cite{vaswani2017attention}, action classification \cite{wang2018non}, image classification \cite{woo2018cbam}, image generation \cite{zhang2019self}, cross-modal translation \cite{duan2019cascade}, text-to-image synthesis \cite{cai2019dualattn}, and crowd counting \cite{chen2020relevant,zhang2019relational}.
For instance, 
Wang et al. \cite{wang2018non} explore the non-local operation in the space-time dimension for video and image processing.
Zhang et al. \cite{zhang2019self} introduce a self-attention mechanism in the generator for image generation.
However, these methods require computing the correlation between every two points in the feature map, leading to expensive computation cost and thus limiting its applications.

Different from the above-mentioned methods, we propose a novel DAGAN for the semantic image synthesis task, and carefully design two modules (i.e., SAM and CAM) to capture semantic attention in both spatial and channel dimensions for improving feature representations. 
Extensive experiments validate the effectiveness of the proposed method.

%% file: 3Method.tex
\section{Dual Attention GANs}
In this section, we first introduce an overview of our method and then present the two attention modules. Finally, we introduce the optimization objective and training details of the proposed whole framework.

\begin{figure*}[!ht] \small
	\centering
	\includegraphics[width=1\linewidth]{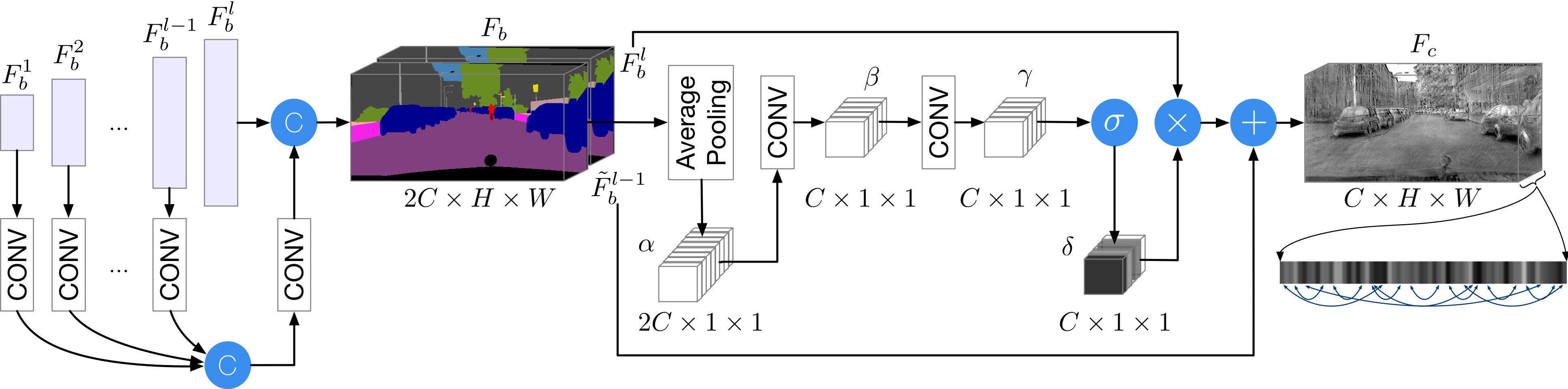}
	\caption{Overview of the proposed scale-wise Channel Attention Module (CAM), which aims to learn the scale-wise attention between channel features with the same object/stuff regardless of their channel distances. The symbols $\oplus$, $\otimes$, $\textcircled{c}$ and $\textcircled{$\sigma$}$ denote element-wise addition, element-wise multiplication, channel-wise concatenation, and Sigmoid($\cdot$) operation, respectively.}
	\label{fig:channel}
	\vspace{-0.2cm}
\end{figure*}

\noindent \textbf{Overview.}
We start by presenting the details of the proposed Dual Attention GANs (DAGAN).
An illustration of the overall framework is shown in Fig.~\ref{fig:method}, which consists of a generator $G$ and discriminator $D$.
The generator $G$ mainly consists of three parts, i.e., a backbone network $B$ extracting deep multi-scale features from the input layout, a position-wise Spatial Attention Module (SAM) modeling the pixel attention in the spatial dimension, and a scale-wise Channel Attention Module (CAM) capturing the feature attention in the channel dimension. 

Intuitively, stuff and objects in the input semantic layout are diverse on scales, lighting, and views. 
The features corresponding to the pixels with the same semantic label may have some differences due to traditional convolution operations that would lead to a local receptive view, resulting in intra-class semantic inconsistency and affect the generation performance (see Fig.~\ref{fig:seg}).
To address this issue, we explore long-range semantic correlations by building attention among spatial pixels and channel features, thus improving feature representation for image generation.
As illustrated in Fig.~\ref{fig:method}, we design two types of attention modules to improve feature representations.

\noindent \textbf{Multi-Scale Feature Extraction.}
We follow previous works \cite{wang2018high,park2019semantic,liu2019learning} and employ the semantic layout $S$ as the input of our backbone network $B$, as shown in Fig.~\ref{fig:method}. 
The network $B$ aims to extract deep multi-scale features of $S$, which can be formulated as, 
\begin{equation}
\begin{aligned}
F_b^i = B(S), \quad {\rm for} \quad i= 1,2,\cdots, l,
\end{aligned}
\end{equation}
where $F_b^i$ denotes the feature map extracted from the $i^{th}$ layer of the backbone network $B$, as shown in Fig.~\ref{fig:method}.
By doing so, we obtain a multi-scale feature representation of $S$ for further processing.

\noindent \textbf{Position-Wise Spatial Attention Modeling.}
Existing methods such as \cite{park2019semantic,liu2019learning,wang2018high} always use local features generated by convolutional operations, leading different generation results of the same label.
To model pixel correlations over local features, we propose a position-wise Spatial Attention Module (SAM), which encodes spatial pixel correlations into local features, enhancing their representation capability. 

The framework of the proposed SAM is elaborated in Fig.~\ref{fig:spatial}.
Specifically, given the local feature $F_b^l {\in} \mathbb{R}^{C\times H \times W}$ extracted from the last layer of the backbone network $B$, we first feed it into average and max pooling operations to produce two new feature maps $P_a {\in} \mathbb{R}^{1\times H \times W}$ and $P_m {\in} \mathbb{R}^{1\times H \times W}$,
\begin{equation}
\begin{aligned}
P_a &= {\rm AvePool}(F_b^l), \\
P_m & = {\rm MaxPool}(F_b^l),
\end{aligned}
\end{equation}
where ${\rm AvePool}(\cdot)$ and ${\rm MaxPool}(\cdot)$ represent average and max pooling, respectively.
Although \cite{cai2019dualattn} and our method both use average and max pooling, the way of using them is different. Specifically, we use both average and max pooling in the spatial dimension since we need to model the correlations between the regions with the same semantic label, while \cite{cai2019dualattn} uses average and max pooling in the channel dimension to enhance the features in the channel dimension.

We then concatenate both $P_a$ and $P_m$ to form a new feature $P_{am} {\in} \mathbb{R}^{2\times H \times W}$. 
After that, we perform a convolutional operation on $P_{am}$, and apply a Sigmoid($\cdot$) activation function to calculate the spatial attention map $A_s{\in} \mathbb{R}^{1\times H \times W}$. Mathematically,
\begin{equation}
\begin{aligned}
A_s = \sigma({\rm Conv}({\rm Concat}(P_a, P_m))),
\end{aligned}
\end{equation}
where $\sigma(\cdot)$, ${\rm Conv}(\cdot)$ and ${\rm Concat}(\cdot)$ denote
Sigmoid function, convolution operation and channel-wise concatenation, respectively.
Next, we perform a matrix multiplication between $A_s$ and the original feature $F_b^l$ to obtain the updated feature $F_s {\in}  \mathbb{R}^{C\times H \times W}$. 
The computation process is summarized as follow,
\begin{equation}
\begin{aligned}
F_s = A_s \otimes F_b^l,
\end{aligned}
\end{equation}
where $\otimes$ denote element-wise multiplication.
Therefore, the spatial attention guided feature $F_s$ has a global contextual view in the spatial dimension.
By doing so, the pixels with the same semantic label achieve mutual gains, thus improving intra-class semantic consistency (see Fig.~\ref{fig:seg}).

\noindent \textbf{Scale-Wise Channel Attention Modeling.}
Each channel map of features can be regarded as a scale-specific response, and the same object/stuff with different semantic responses should be correlated and associated with each other. 
To exploit the correlations between channel maps for enhancing the consistency, we propose a scale-wise Channel Attention Module (CAM) to explicitly reason the scale-wise correlations between channels.

The structure of the proposed CAM is illustrated in Fig.~\ref{fig:channel}. 
Different from SAM, we first reshape $\{F_b^i\}_{i=1}^{l-1}$ to the same size of $F_B^l$ and then feed them to convolution layers.
Next, we concatenate all of them and feed the result into a new convolution layer to obtain a new feature $\tilde{F}_b^{l-1} {\in} \mathbb{R}^{C\times H \times W}$.
This process can expressed as,
\begin{equation}
\begin{aligned}
\tilde{F}_b^{l-1} = {\rm Conv}({\rm Concat}({\rm Conv}(F_b^1), {\rm Conv}(F_b^2), \cdots, {\rm Conv}(F_b^{l-1}))),
\end{aligned}
\end{equation}
where ${\rm Conv}(\cdot)$ and ${\rm Concat}(\cdot)$ denote convolution operation and channel-wise concatenation, respectively.
After that, we concatenate $\tilde{F}_b^{l-1}$ and $F_b^l$, and feed the result to an average pooling layer to obtain a scale vector $\alpha {\in} \mathbb{R}^{2C\times 1 \times 1}$. Mathematically
\begin{equation}
\begin{aligned}
\alpha = {\rm AvePool}({\rm Concat}(\tilde{F}_b^{l-1}, F_b^l)),
\end{aligned}
\end{equation}
where ${\rm AvePool}(\cdot)$ denote the average pooling.
To reduce the number of the channel of $\alpha$, we feed it to two successively convolution layers to obtain a new scale vector $\gamma{\in} \mathbb{R}^{C\times 1 \times 1}$, indicating the weights of different channels are equal.
 
However, the features in different scales have different degrees of discrimination, which leads to different consistency of generation. 
To obtain the intra-class consistent generation, we extract the discriminative features within the same label and inhibit the indiscriminative features between different labels. 
Specifically, we apply a Sigmoid($\cdot$) activation function to obtain the channel attention weight $\delta {\in} \mathbb{R}^{C\times 1 \times 1}$.
This process can be formulated as,
\begin{equation}
\begin{aligned}
\delta = \sigma(\gamma).
\end{aligned}
\end{equation}
By doing so, each item in the channel attention weight $\delta$ measures the importance of the corresponding channel.
Finally, we introduce two ways to calculate the updated feature $F_c {\in} \mathbb{R}^{C\times H \times W}$, which represents the feature selection with CAM.
The first one is using the channel weight $\delta$ to multiply $\tilde{F}_b^{l-1}$ and perform an element-wise sum with $F_b^l$,
\begin{equation}
\begin{aligned}
F_c = \delta \otimes \tilde{F}_b^{l-1} + F_b^l,
\end{aligned}
\label{eq:method1}
\end{equation}
where $\otimes$ denotes element-wise multiplication.
The second one is to use the channel weight $\delta$ to multiply $F_b^l$ and perform an element-wise sum with $\tilde{F}_b^{l-1}$,
\begin{equation}
\begin{aligned}
F_c = \delta \otimes F_b^l + \tilde{F}_b^{l-1}.
\end{aligned}
\label{eq:method2}
\end{equation}
In this way, the new feature $F_c$ spotlights attention within the same category regardless of their scales and channel distances, boosting feature discriminability.

Note that the proposed CAM is designed to change the weights of the features on each scale to enhance the scale consistency. With this design, we can make the generator to obtain scale-wise discriminative features, making the generated image to be intra-class consistent. Oppositely, the channel attention module proposed in~\cite{cai2019dualattn} did not consider the multi-scale features causing as such the generated image to be intra-class inconsistent.

\begin{figure*}[!t] \small
	\centering
	\includegraphics[width=0.9\linewidth]{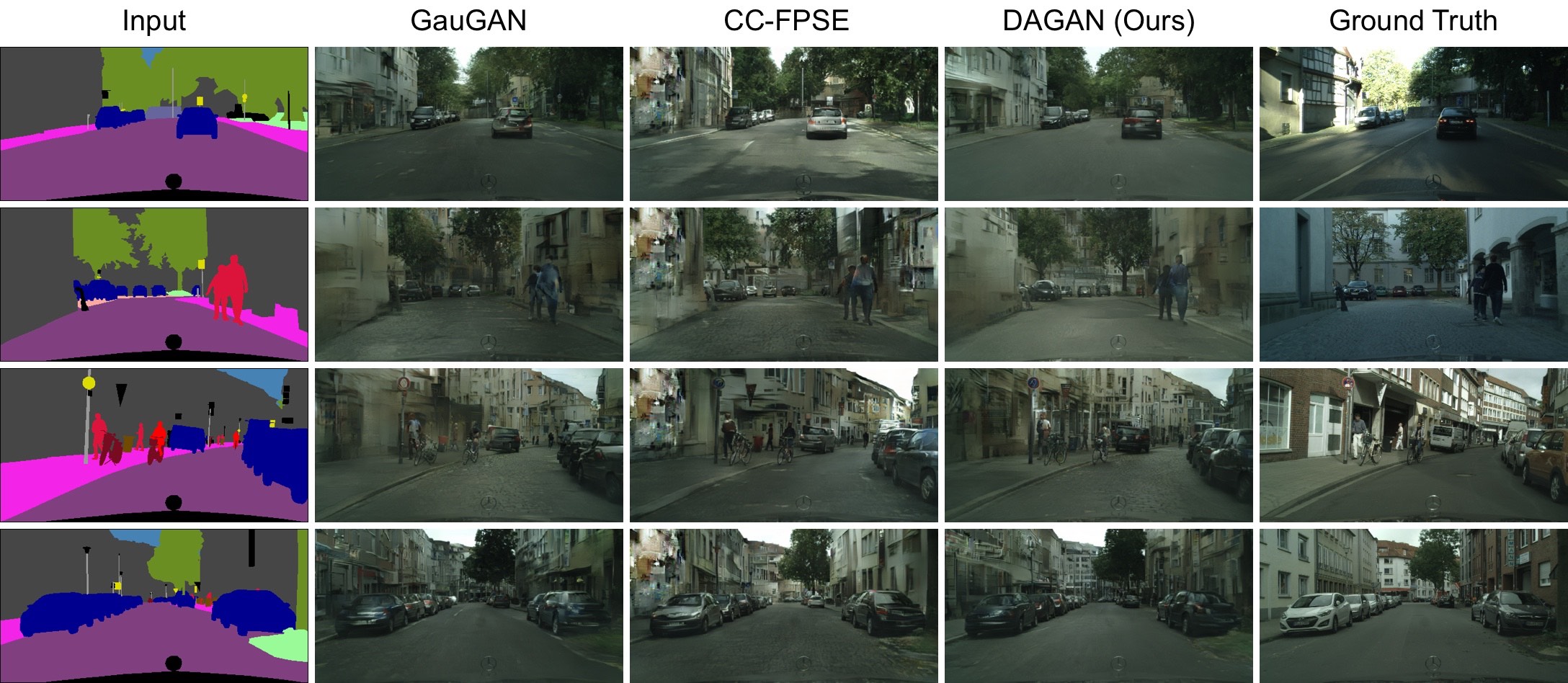}
	\caption{Qualitative comparison on Cityscapes. From left to right: Input, GauGAN~\cite{park2019semantic}, CC-FPSE~\cite{liu2019learning}, DAGAN (Ours) and GT.
	}
	\label{fig:city_results}
	\vspace{-0.2cm}
\end{figure*}

\begin{figure*}[!t] \small
	\centering
	\includegraphics[width=0.9\linewidth]{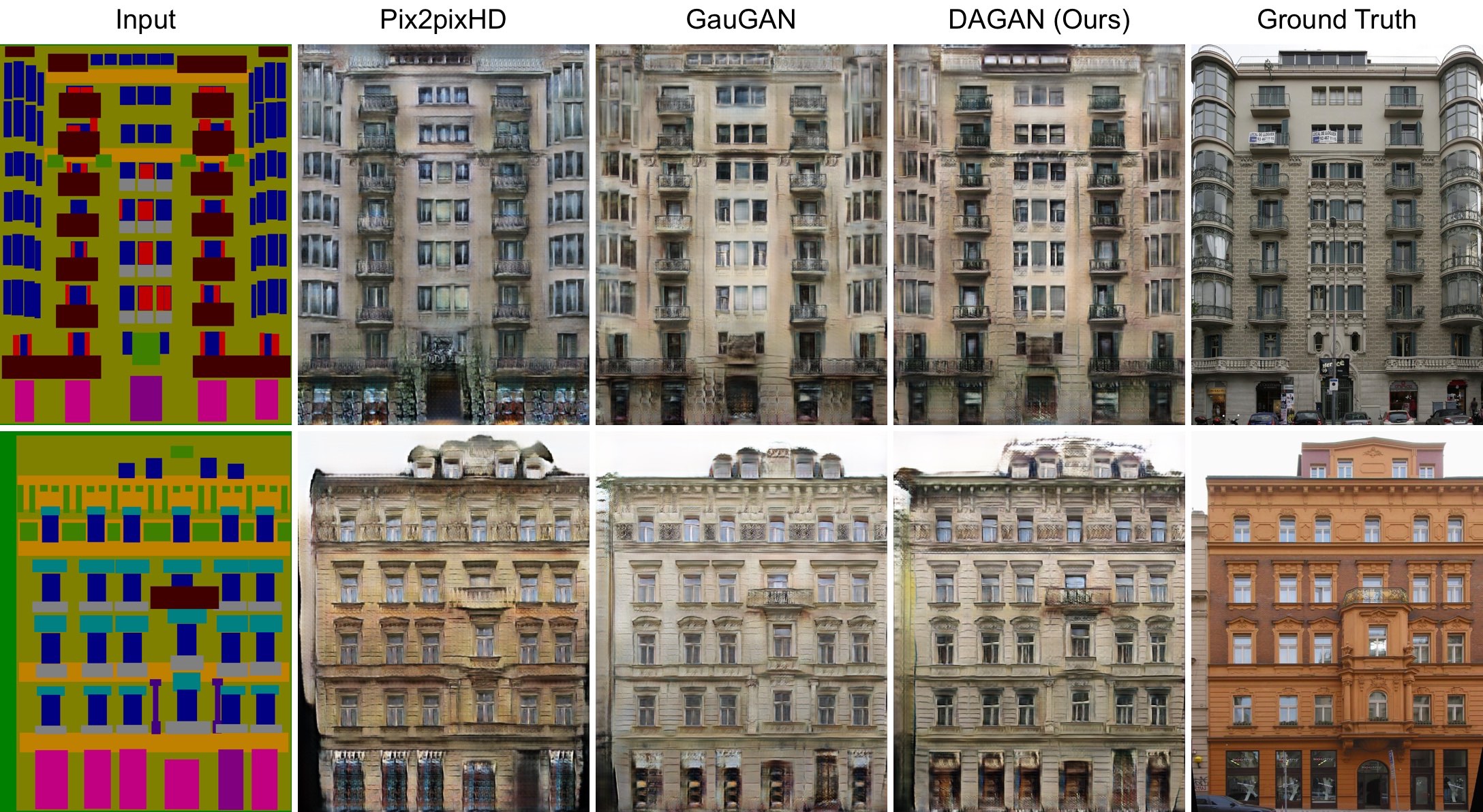}
	\caption{Qualitative comparison on Facades. From left to right: Input, Pix2PixHD~\cite{wang2018high}, GauGAN~\cite{park2019semantic}, DAGAN (Ours) and GT.}
	\label{fig:facades_results}
	\vspace{-0.2cm}
\end{figure*}

\noindent \textbf{Attention Modeling with GANs.}
To take full advantage of pixel and feature attention in both spatial and channel dimensions, we sum the outputs from the two attention modules to obtain better feature representations for image generation.
At last, we adopt a convolution layer to generate the final result $I^{'}$, as shown in Fig.~\ref{fig:method}. 
Notably, the proposed attention modules are simple and can be directly inserted in the existing GAN frameworks without introducing too many parameters computational costs.

\noindent \textbf{Optimization Objective.}
We follow \cite{park2019semantic,liu2019learning} and employ three different losses as our optimization objective.
\begin{equation}
\begin{aligned}
\mathcal{L} =  \lambda_{cgan} \mathcal{L}_{cgan} + \lambda_{f} \mathcal{L}_{f} + \lambda_{p} \mathcal{L}_{p},
\label{eq:loss} 
\end{aligned}
\end{equation}
where $\mathcal{L}_{cgan}$, $\mathcal{L}_{f}$ and $\mathcal{L}_{p}$ denote the conditional adversarial loss, the discriminator feature matching loss and the perceptual loss, respectively.
We set $\lambda_{cgan}{=}1$, $\lambda_{f}{=}10$ and $\lambda_{p}{=}10$ in our experiments.

\noindent \textbf{Training Details.}
We employ the multi-scale discriminator used in \cite{wang2018high,park2019semantic} as our discriminator $D$.
We follow the training procedures of GANs and alternatively train the generator $G$ and discriminator $D$, i.e., one gradient descent step on discriminator and generator alternately. 
We use the Adam solver \cite{kingma2014adam} and set $\beta_1{=}0$, $\beta_2{=}0.999$.
We conduct the experiments on NVIDIA DGX1 with 8 32GB V100 GPUs

%% file: 4Experiments.tex
\section{Experiments}

\noindent \textbf{Datasets.}
We conduct extensive experiments on four public datasets to validate the proposed DAGAN, i.e., Cityscapes \cite{cordts2016cityscapes}, ADE20K \cite{zhou2017scene}, CelebAMask-HQ \cite{CelebAMask-HQ} and Facades \cite{tylevcek2013spatial}.
Notably, we follow the same train/test split used in their papers.
Moreover, to verify the robustness of the proposed DAGAN on different image resolutions, we resize the images to $256 {\times} 256$, $512 {\times} 256$, $512 {\times} 512$, and $1024 {\times} 1024$ on ADE20K, Cityscapes, CelebAMask-HQ, and Facades, respectively.

\begin{figure*}[!t]\small
	\centering
	\includegraphics[width=0.9\linewidth]{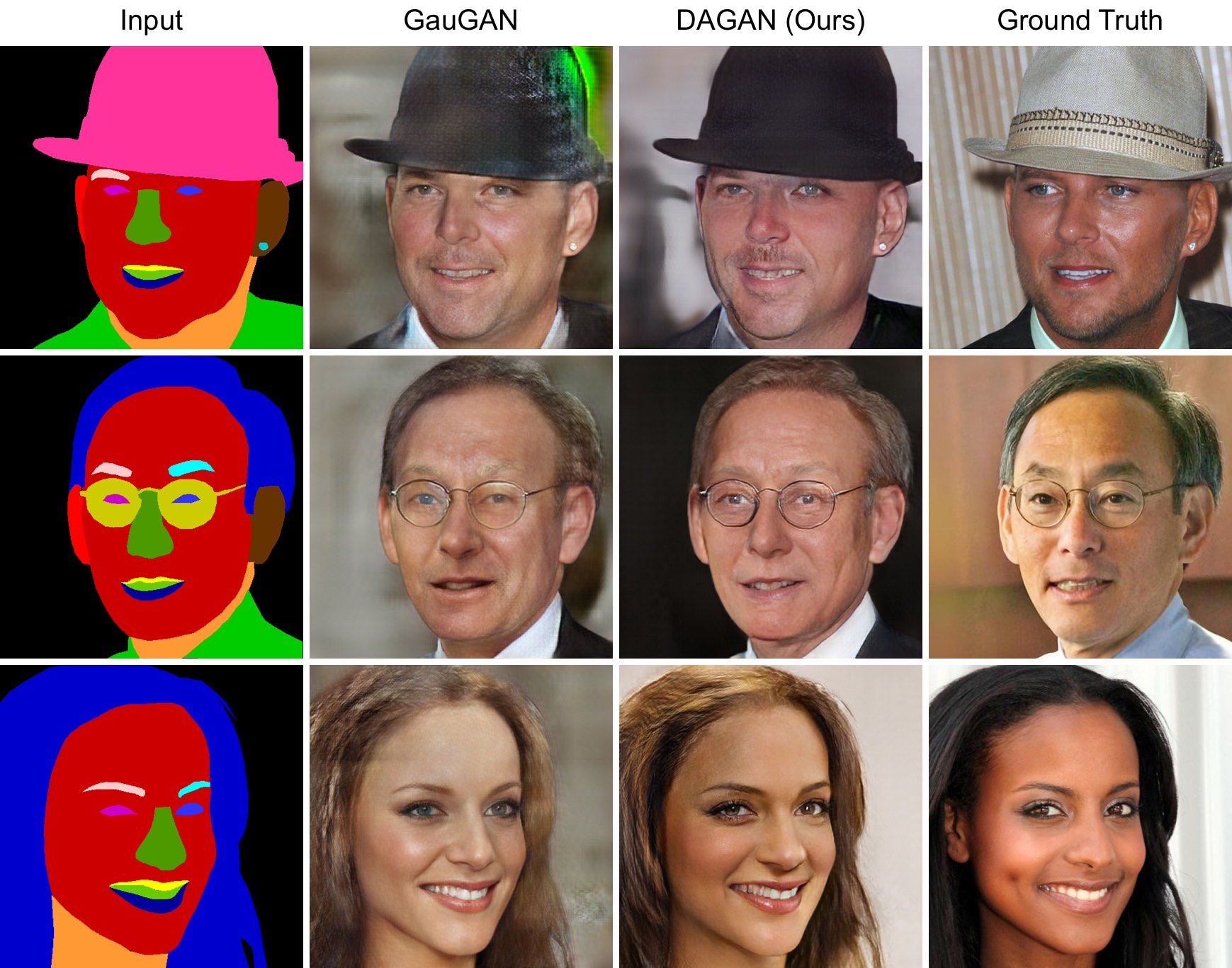}
	\caption{Qualitative comparison on CelebAMask-HQ. From left to right: Input, GauGAN~\cite{park2019semantic}, DAGAN (Ours) and GT.
	}
	\label{fig:celeba}
	\vspace{-0.2cm}
\end{figure*}

\noindent \textbf{Evaluation Metrics.}
We follow GauGAN~\cite{park2019semantic} and use segmentation accuracy, i.e., mean Intersection-over-Union (mIoU) and pixel accuracy (Acc), as our evaluation metrics.
Moreover, we use Fr\'echet Inception Distance (FID)~\cite{heusel2017gans}, Fr\'echet ResNet Distance (FRD) \cite{tang2018gesturegan} and Learned Perceptual Image Patch Similarity (LPIPS) \cite{zhang2018unreasonable} to evaluate the feature distance between generated images and real samples.

\subsection{Comparisons with State-of-the-Art}
\begin{figure*}[!t] \small
	\centering
	\includegraphics[width=0.92\linewidth]{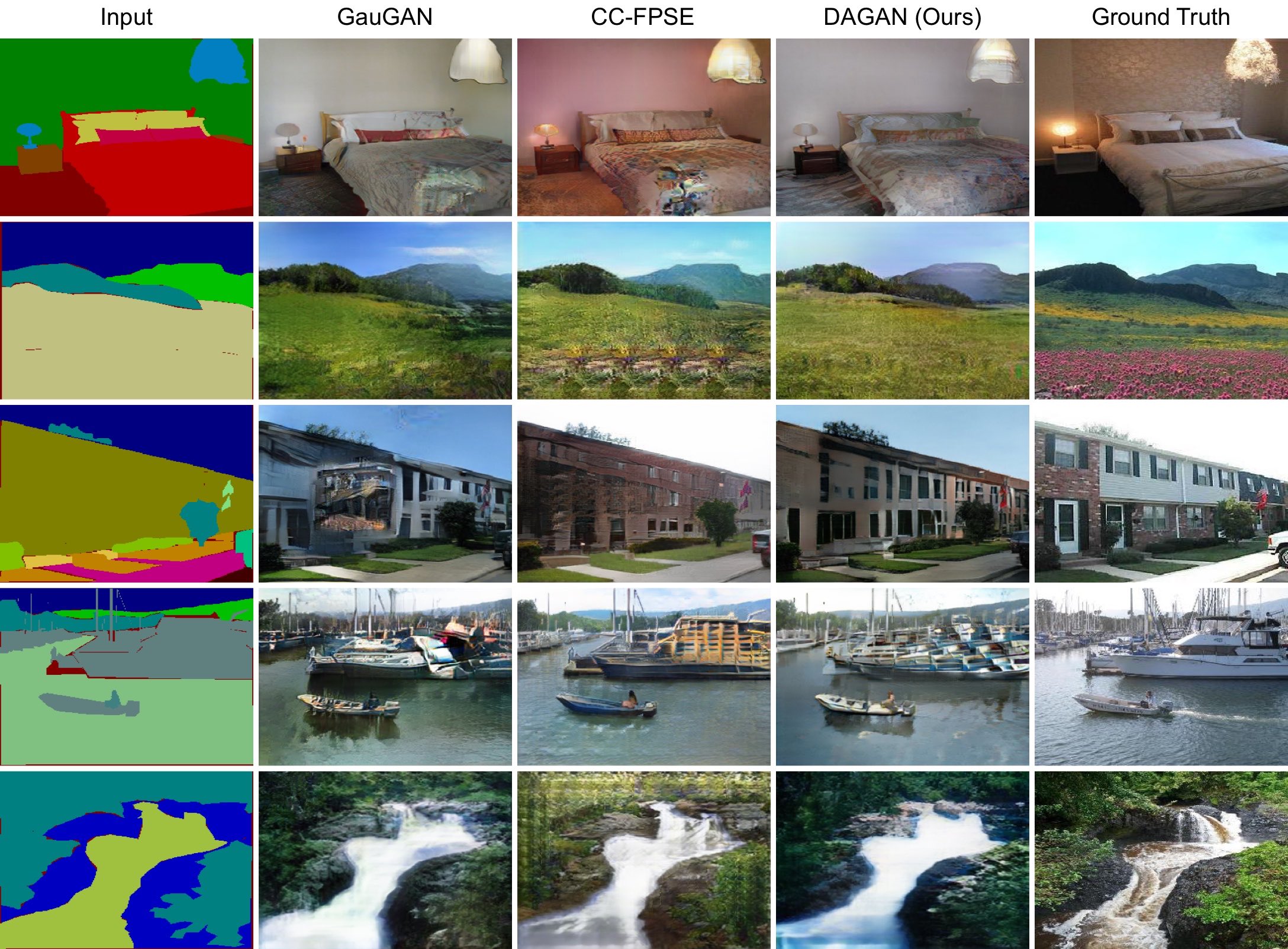}
	\caption{Qualitative comparison on ADE20K. From left to right: Input, GauGAN~\cite{park2019semantic}, CC-FPSE~\cite{liu2019learning}, DAGAN (Ours) and GT.}
	\label{fig:ade_results}
	\vspace{-0.2cm}
\end{figure*}

\noindent \textbf{Qualitative Comparisons.}
We compare the proposed DAGAN with several leading methods, i.e., GauGAN \cite{park2019semantic}, Pix2pixHD \cite{wang2018high} and CC-FPSE \cite{liu2019learning}.
Comparison results are shown in Fig.~\ref{fig:city_results}, \ref{fig:facades_results}, \ref{fig:celeba} and \ref{fig:ade_results}.
We observe that the proposed method generates more clear and visually plausible images than the existing baselines, validating the effectiveness of the proposed DAGAN.
Note that we cannot reproduce the results of CC-FPSE on both Facades and CelebAMask-HQ datasets because we cannot fit the CC-FPSE model to our GPUs on both datasets.

\begin{table}[!t]\small
	\centering
	\caption{User study. 
	The numbers indicate the percentage of users who favor the results of the proposed DAGAN over the competing method. 
	For this metric, higher is better.
	}
		\resizebox{1\linewidth}{!}{%
	\begin{tabular}{lcccc} \toprule
		AMT $\uparrow$                                     & Cityscapes & ADE20K & Facades & CelebAMask-HQ \\ \midrule
		Ours vs. GauGAN~\cite{park2019semantic}  & 60.71  & 64.32 & 63.17 & 67.92 \\ 
		Ours vs. CC-FPSE~\cite{liu2019learning}   &  57.38 & 59.39 & - & - \\ \bottomrule
	\end{tabular}}
	\label{tab:amt}
	\vspace{-0.2cm}
\end{table}

\noindent \textbf{User Study.}
We follow the evaluation protocol of GauGAN \cite{park2019semantic} and perform a user study to measure the quality of generated images.
Comparison results are shown in Table~\ref{tab:amt}.
We observe that users strongly favor the images generated by the proposed DAGAN than both GauGAN and CC-FPSE on all the four challenging datasets, further validating that the generated images by our method are more photo-realistic.

\begin{table*}[!t]\small
	\centering
	\caption{Quantitative comparison of different methods. For all metrics except mIoU and Acc, lower is better.}
	\begin{tabular}{lcccccccccccc} \toprule
		\multirow{2}{*}{Method}  & \multicolumn{3}{c}{Cityscapes} & \multicolumn{3}{c}{ADE20K} & \multicolumn{3}{c}{Facades} & \multicolumn{3}{c}{CelebAMask-HQ}\\ \cmidrule(lr){2-4} \cmidrule(lr){5-7}  \cmidrule(lr){8-10} 
		\cmidrule(lr){11-13} 
		& mIoU $\uparrow$ & Acc $\uparrow$  & FID  $\downarrow$ & mIoU $\uparrow$    & Acc $\uparrow$  & FID  $\downarrow$  & FID $\downarrow$ & LPIPS $\downarrow$ & FRD $\downarrow$ & FID $\downarrow$ & LPIPS $\downarrow$ & FRD $\downarrow$\\ \midrule
		CRN~\cite{chen2017photographic} & 52.4  & 77.1 & 104.7  & 22.4 & 68.8 & 73.3 & - & - & - & - & - & - \\
		SIMS~\cite{qi2018semi}  & 47.2  & 75.5 & \textbf{49.7} & -  & - & - & - & - & - & - & - & - \\
		Pix2pixHD~\cite{wang2018high}    & 58.3  & 81.4 & 95.0  & 20.3 & 69.2 & 81.8 & 128.5 & 0.6466 & 3.7402 & - & - & - \\
		GauGAN~\cite{park2019semantic} & 62.3  & 81.9 & 71.8  & 38.5 & 79.9 & 33.9 & 127.2 & 0.6268 & 3.5309 & 42.2 & 0.4870 & \textbf{3.4523}\\
		CC-FPSE~\cite{liu2019learning}  & 65.5 & 82.3 & 54.3 & \textbf{43.7} & \textbf{82.9} & \textbf{31.7} & - & - & - & - & - & - \\
		DAGAN (Ours)  & \textbf{66.1}  & \textbf{82.6} & 60.3 & 40.5 & 81.6 & 31.9 & \textbf{116.6} & \textbf{0.6224} & \textbf{3.4929} & \textbf{23.9} & \textbf{0.4796} & 3.4562 \\	\bottomrule
	\end{tabular}
	\vspace{-0.2cm}
	\label{tab:sota}
\end{table*}

\begin{table*}[!t]\small
	\centering 
	\caption{Quantitative comparison of model parameters. `Gen.' and  `Dis.' denote Generator and Discriminator, respectively.}
	\begin{tabular}{lcccccccccccc} \toprule
		\multirow{2}{*}{Method}   & \multicolumn{3}{c}{Cityscapes} & \multicolumn{3}{c}{ADE20K} & \multicolumn{3}{c}{Facades} & \multicolumn{3}{c}{CelebAMask-HQ} \\ \cmidrule(lr){2-4} \cmidrule(lr){5-7} \cmidrule(lr){8-10} 
		\cmidrule(lr){11-13}
		& Gen.  & Dis. & Total $\downarrow$   & Gen. & Dis. & Total $\downarrow$  & Gen. & Dis. & Total $\downarrow$ & Gen. & Dis. & Total $\downarrow$  \\ \midrule
		GauGAN~\cite{park2019semantic}   &  93.0M & 5.6M & \textbf{98.6M} & 96.5M & 5.8M & \textbf{102.3M} & 92.4M  & 5.6M & \textbf{98.0M} & 92.5M & 5.6M & \textbf{98.1M} \\
		CC-FPSE~\cite{liu2019learning}  & 138.6M & 5.2M & 143.8M & 151.2M & 5.2M & 156.4M  & 398.7M & 5.2M & 403.9M & 196.8M & 5.2M & 202.0M \\
		DAGAN (Ours) & 93.1M & 5.6M & 98.7M & 96.6M & 5.8M & 102.4M & 92.4M & 5.6M & \textbf{98.0M} & 92.6M & 5.6M & 98.2M \\	\bottomrule
	\end{tabular}
	\label{tab:para}
	\vspace{-0.2cm}
\end{table*}

\noindent \textbf{Quantitative Comparisons.} 
We also provide quantitative results in Table \ref{tab:sota}.
Clearly, the proposed DAGAN achieves the best results compared with the baselines except CC-FPSE \cite{liu2019learning} on ADE20K.
However, we see that the proposed DAGAN generates more photo-realistic images with fewer artifacts than CC-FPSE in Fig.~\ref{fig:ade_results}.
Moreover, we provide the number of model parameters in Table~\ref{tab:para}.
We see that the proposed DAGAN has remarkably fewer model parameters than CC-FPSE, which means DAGAN requires significantly less training time and GPU memory than CC-FPSE.
Notably, for CC-FPSE, we cannot generate high-resolution images on both Facades ($1024 {\times} 1024$) and CelebAMask-HQ ($512 {\times} 512$) datasets since CC-FPSE has many parameters that need to be learned on both datasets, resulting in GPU memory overflow.

\begin{figure*}[!t] \small
	\centering
	\includegraphics[width=0.92\linewidth]{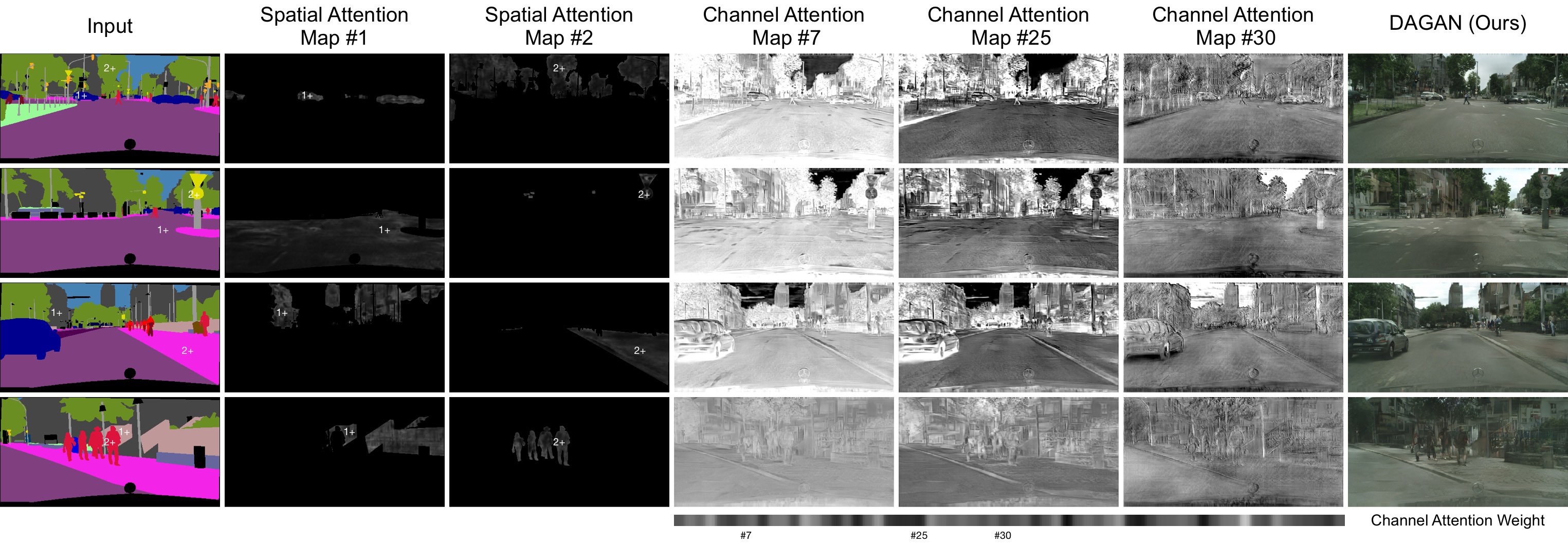}
	\caption{Visualization of learned spatial and channel attention maps on Cityscapes.
	}
	\label{fig:att}
	\vspace{-0.2cm}
\end{figure*}


\noindent \textbf{Visualization of Learned Attention Maps.}
For spatial attention map, we randomly select two classes on each sample and display their corresponding spatial attention map in columns 2 and 3 in Fig.~\ref{fig:att}, respectively. 
We see that the spatial attention module captures global relationships within each semantic class. 
For instance, in the first row, the point `1+' is marked on a car and its spatial attention map (in column 2) highlights most the areas where the cars are. 
For channel attention map, we show the $7^{th}$, $25^{th}$ and $30^{th}$ channel attention map in column 4, 5 and 6 in Fig.~\ref{fig:att}, respectively. 
We see that the difference of each channel is noticeable after going through the proposed channel attention module. 
For example, the $7^{th}$ channel map is brighter than both $25^{th}$ and $30^{th}$ channel maps, which means the $7^{th}$ channel map contains more information and is more important than both $25^{th}$ and $30^{th}$ channel maps.
Moreover, we show the learned channel attention wight of all channels, which indicates the same brightness has the same importance regardless of object scales and channel distances.



\noindent \textbf{Visualization of Generated Semantic Maps.}
We follow GauGAN \cite{park2019semantic} and apply pretrained semantic segmentation models on the generated images to produce semantic maps.
Specifically, we employ DRN-D-105 \cite{yu2017dilated} for Cityscapes and UperNet101 \cite{xiao2018unified} for ADE20K.
The generated semantic maps are shown in Fig. \ref{fig:seg}.
We see that the proposed DAGAN generates more intra-class semantic consistency labels than GauGAN, confirming our initial motivation.

\subsection{Ablation Study}
We conduct extensive ablation studies on Cityscapes \cite{cordts2016cityscapes} to evaluate each component of the proposed DAGAN.

\noindent \textbf{Baselines of DAGAN.}
The proposed DAGAN has six baselines (i.e., B1, B2, B3, B4, B5, B6) as shown in Table \ref{tab:abla}.
(i) B1 is our baseline. 
(ii) B2 uses the proposed Spatial Attention Module (SAM) to model the position-wise pixel attention in the spatial dimension.
(iii) B3 employ the proposed Channel Attention Module (CAM) to reason the scale-wise feature attention in the channel dimension. Note that it uses Eq. \eqref{eq:method1} to calculate the new channel feature $F_c$. 
(iv) The difference between B4 and B3 is that B4 uses Eq. \eqref{eq:method2} to calculate the new channel feature $F_c$.
(v) B5 combines both SAM and CAM-I to model the semantic attention in both spatial and channel dimensions.
(vi) B6 is our final model and adopts the combination of SAM and CAM-II to reason both spatial and channel semantic attentions.

\noindent \textbf{Ablation Analysis.}
The results of ablation study are shown in Table \ref{tab:abla}. 
By comparison B2 with B1, the proposed SAM improves mIoU, Acc and FID by 2.9, 0.7 and 7.3, respectively, which confirms the importance of modeling the position-wise spatial pixel attention.
By using the proposed CAM-I, B3 outperforms B1 on mIoU, Acc and FID by 3.3, 0.9 and 8.3, respectively, confirming the effectiveness of the proposed CAM.
B4 outperforms B3 showing CAM-II is more effective than CAM-I.
B5 significantly outperforms both B2 and B3, demonstrating the effectiveness of modeling both spatial and channel semantic attentions for generating photo-realistic and semantically-consistent images.
Finally, we observe that by combining both SAM and CAM-II, the overall performance is further boosted, demonstrating the advantage of our full model.

\noindent \textbf{Comparisons with \cite{cai2019dualattn}.}
Lastly, we compare the proposed method with~\cite{cai2019dualattn} on Cityscapes. 
Specifically, we use the visual attention module proposed in \cite{cai2019dualattn} to replace the dual-attention module proposed in our DAGAN, obtaining the following results in terms of mIoU, Acc, and FID: 64.8, 82.2, and 63.8, respectively. We can see that our method still significantly outperforms \cite{cai2019dualattn}.

\begin{table}[!t] \small
	\centering
	\caption{Ablation study of our DAGAN on Cityscapes. For all metrics except FID, higher is better. `SAM' and `CAM' represents the proposed position-wise Spatial and scale-wise Channel Attention Module, respectively.}
	\begin{tabular}{clccc} \toprule
		& Settings  &  mIoU $\uparrow$ & Acc $\uparrow$ & FID $\downarrow$  \\ \midrule	
		B1 & Baseline                    &  61.3     & 81.5   & 71.8 \\
		B2 & B1 + SAM                 &  64.2     & 82.2  & 64.5 \\
		B3 & B1 + CAM-I              &  64.6     & 82.4  & 63.5 \\
		B4 & B1 + CAM-II             &  65.6     & 82.4  & 62.8 \\
		B5 & B2 + CAM-I   & 65.8    & 82.6   & 60.2  \\
		B6 & B2 + CAM-II  &  \textbf{66.1}     & \textbf{82.6}   & \textbf{60.3}  \\  \bottomrule
	\end{tabular}
	\label{tab:abla}
	\vspace{-0.2cm}
\end{table}

%% file: 5Conclusion.tex
\section{Conclusions}
We propose a novel Dual Attention GAN
(DAGAN) for the challenging semantic image synthesis task.
Specifically, we present two new modules, i.e., SAM and CAM. 
SAM is used to model position-wise pixel attention in spatial dimension. 
CAM is used to reason scale-wise feature attention in channel dimension.
The outputs of SAM and CAM are combined to further improve feature representation. 
Experiments on four datasets show that DAGAN achieves remarkably better results than existing methods.
Moreover, both SAM and CAM are lightweight and general modules, and can be seamlessly integrated into any existing GAN-based architectures to strengthen feature representation with negligible overheads.

\begin{acks}
This work has been partially supported by the Italy-China collaboration project TALENT.
\end{acks}

%% file: supplementary.tex
This document provides additional experimental results on the semantic image synthesis task. 
First, we compare the proposed DAGAN with state-of-the-art methods, i.e., Pix2pixHD \cite{wang2018high}, GauGAN \cite{park2019semantic} and CC-FPSE \cite{liu2019learning} (Sec. \ref{sec:s1}). 
Additionally, we show some learned attention maps of the proposed DAGAN (Sec. \ref{sec:s2}).
Finally, we provide the visualization results of the generated semantic maps (Sec. \ref{sec:s3}). 

\section{State-of-the-art Comparison}
\label{sec:s1}
\begin{figure*} [!t]
	\centering
	\includegraphics[width=1\linewidth]{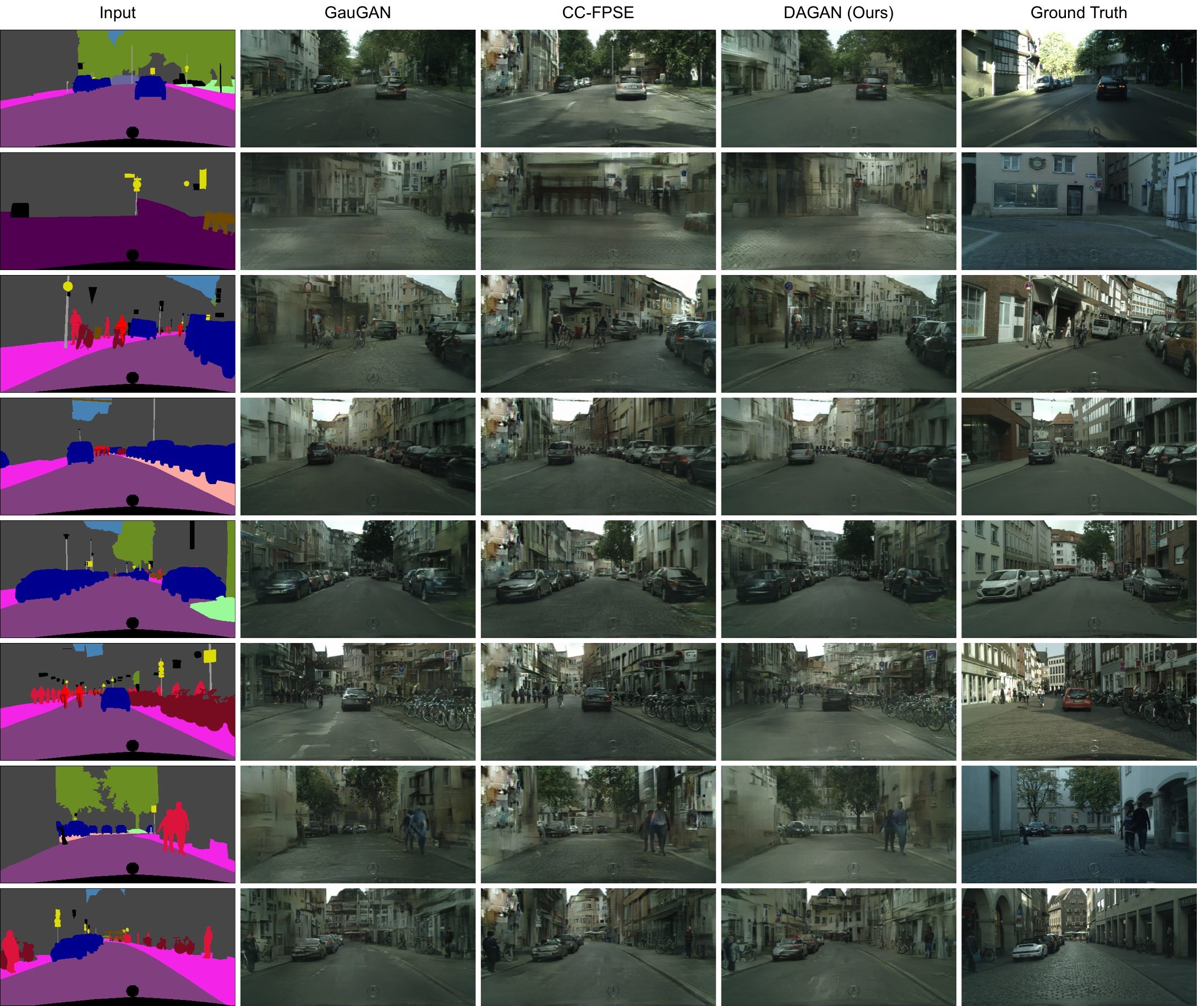}
	\caption{Qualitative comparison on Cityscapes. From left to right: Input, GauGAN~\cite{park2019semantic}, CC-FPSE~\cite{liu2019learning}, DAGAN (Ours) and GT. These samples were randomly selected without cherry-picking for visualization purposes.
	}
	\label{fig:supp_city_results}
\end{figure*}

\begin{figure*} [!t]
	\centering
	\includegraphics[width=1\linewidth]{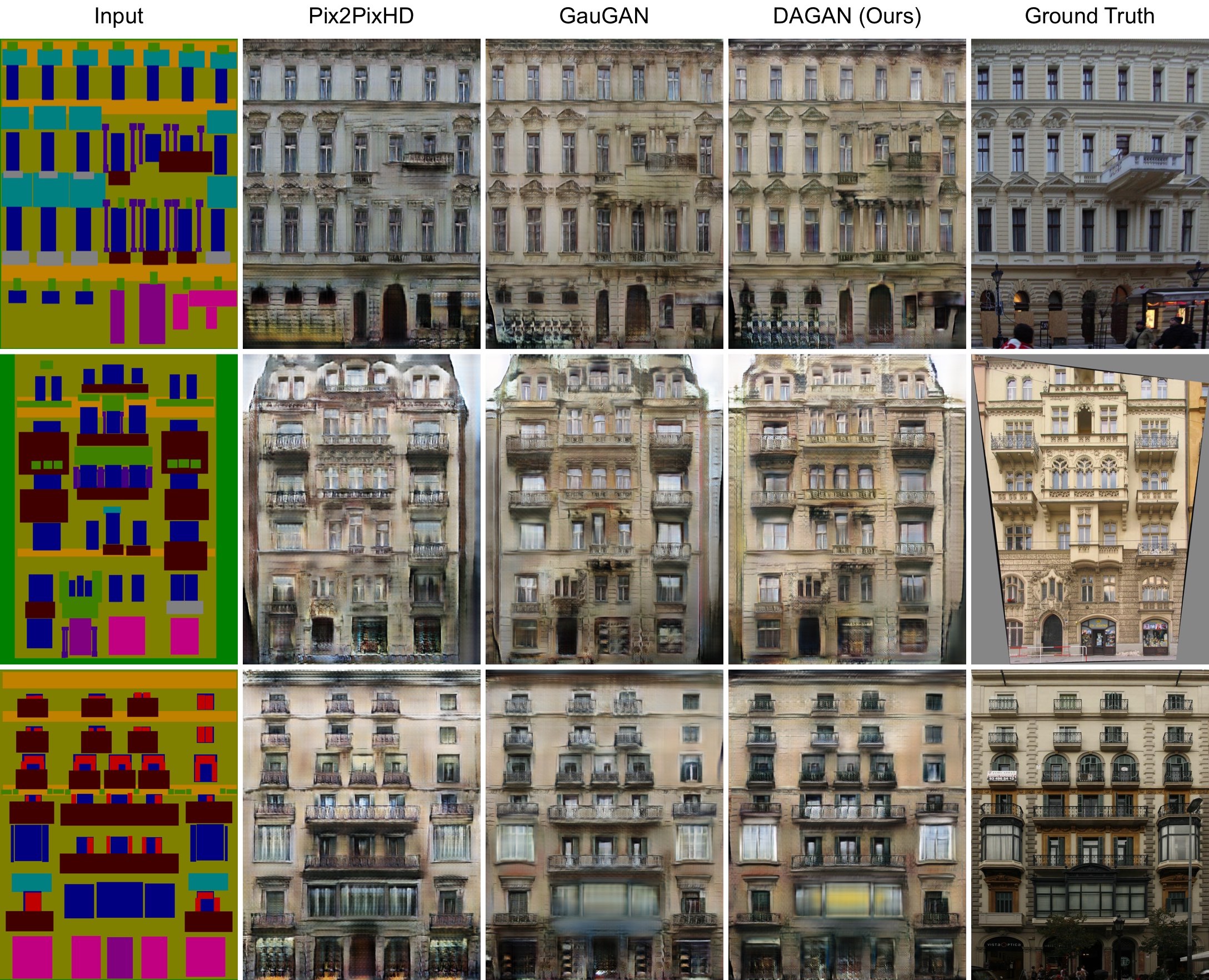}
	\caption{Qualitative comparison on Facades. From left to right: Input, Pix2PixHD~\cite{wang2018high}, GauGAN~\cite{park2019semantic}, DAGAN (Ours) and GT. These samples were randomly selected without cherry-picking for visualization purposes.}
	\label{fig:supp_facades_results}
\end{figure*}

\begin{figure*}[!t]
	\centering
	\includegraphics[width=1\linewidth]{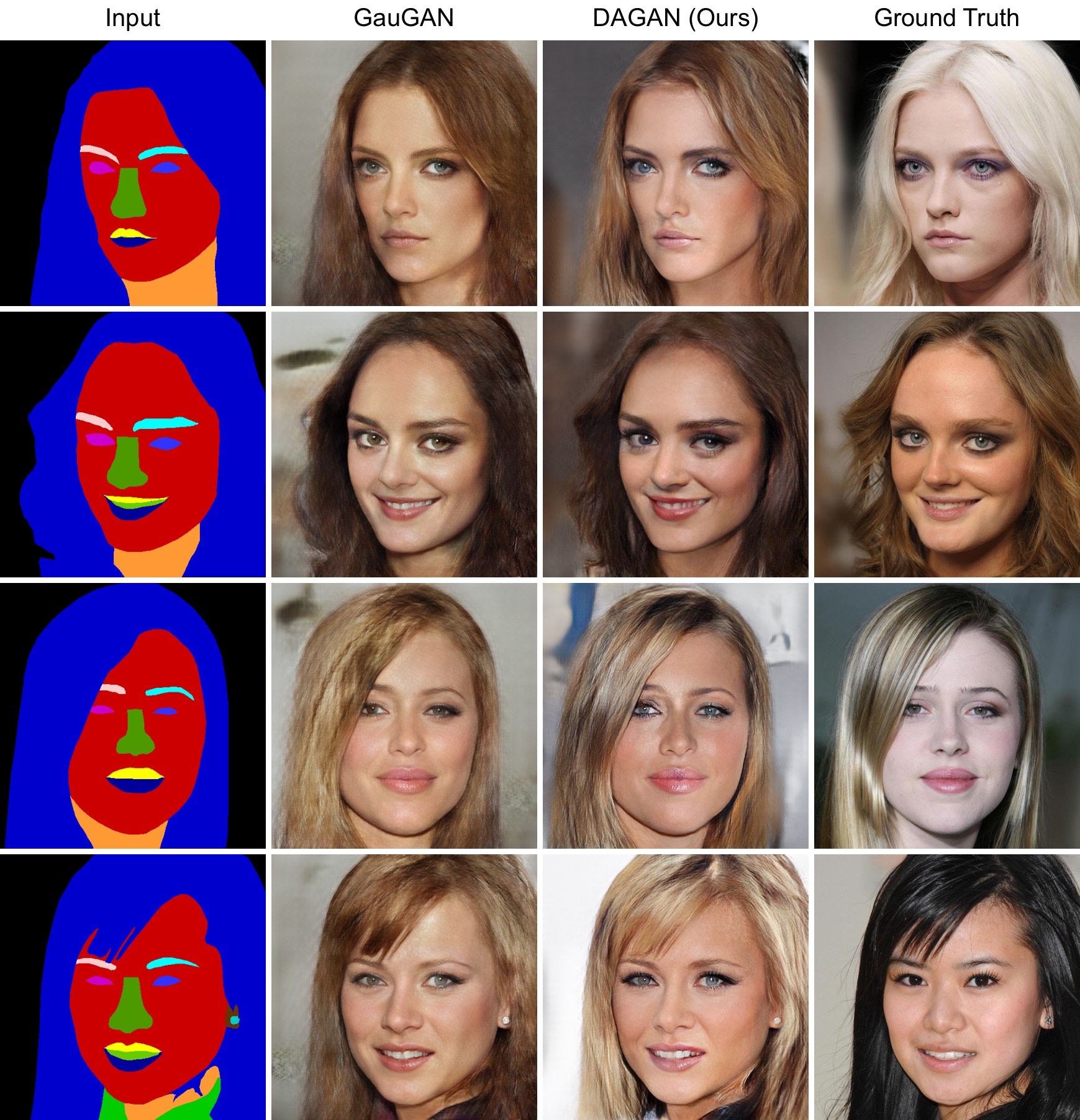}
	\caption{Qualitative comparison on CelebAMask-HQ. From left to right: Input, GauGAN~\cite{park2019semantic}, DAGAN (Ours) and GT. These samples were randomly selected without cherry-picking for visualization purposes.
	}
	\label{fig:celeba_results_supp1}
\end{figure*}

\begin{figure*}[!t]
	\centering
	\includegraphics[width=1\linewidth]{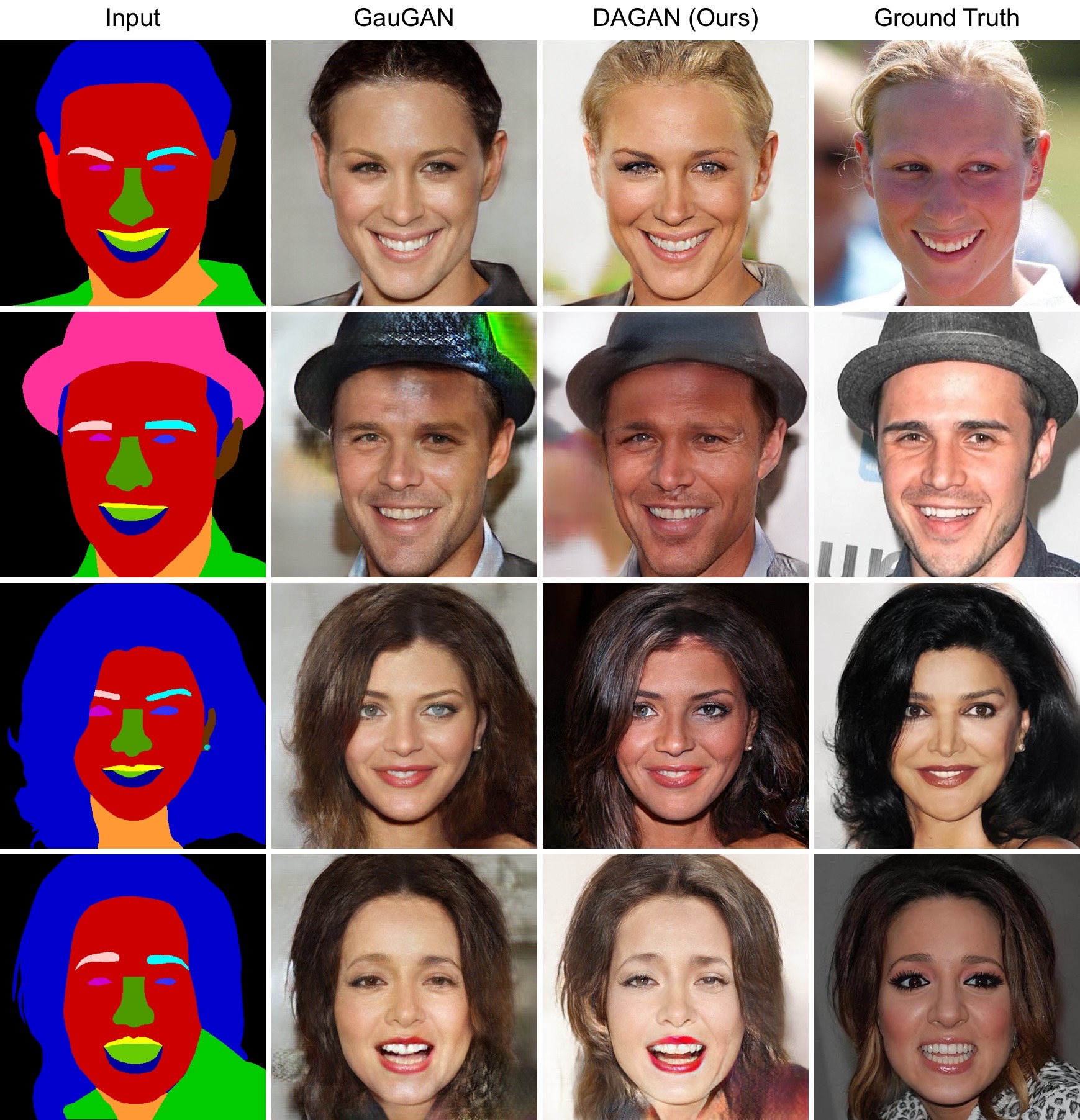}
	\caption{Qualitative comparison on CelebAMask-HQ. From left to right: Input, GauGAN~\cite{park2019semantic}, DAGAN (Ours) and GT. These samples were randomly selected without cherry-picking for visualization purposes.
	}
	\label{fig:celeba_results_supp2}
\end{figure*}

\begin{figure*} [!t]
	\centering
	\includegraphics[width=1\linewidth]{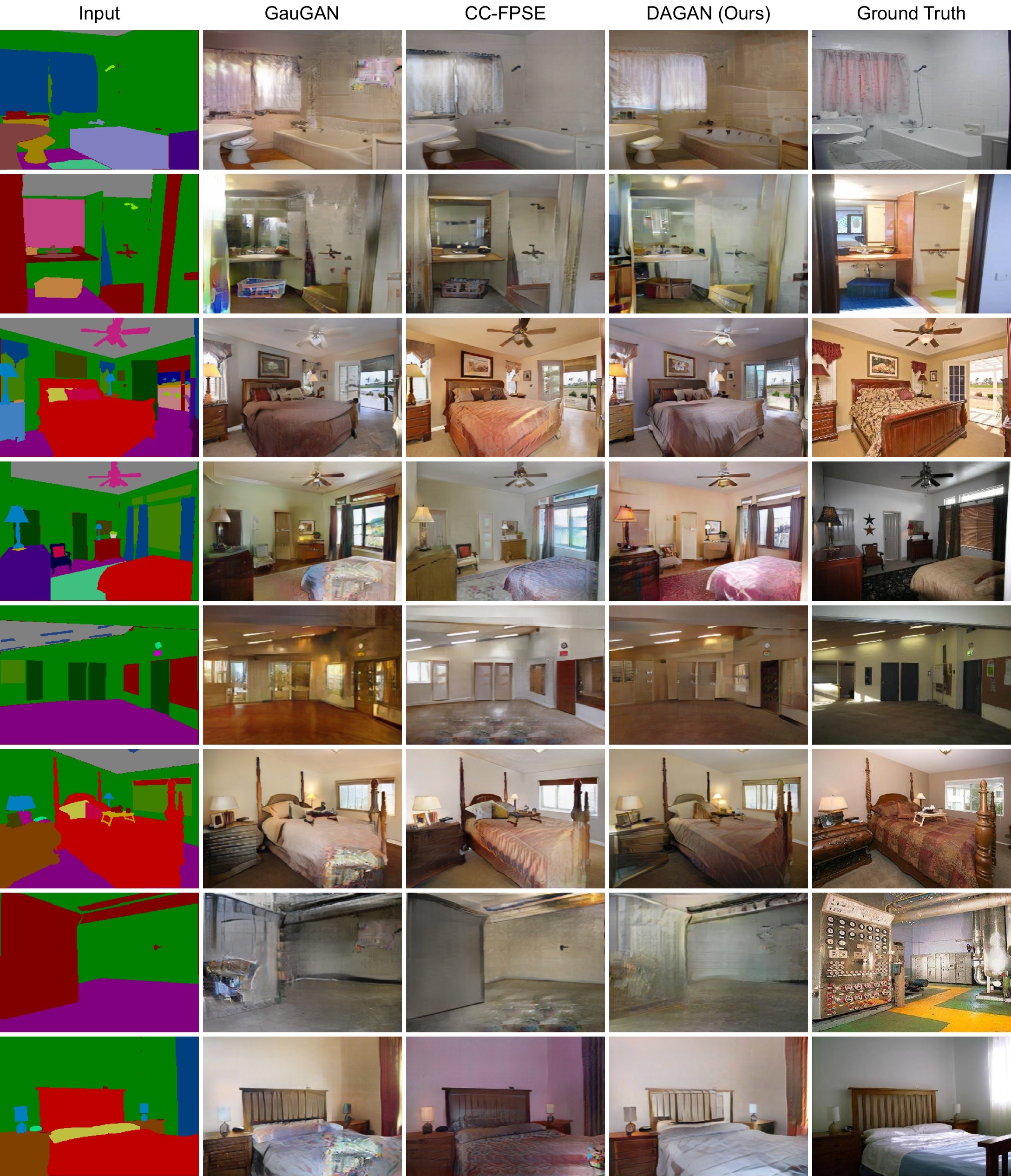}
	\caption{Qualitative comparison on ADE20K. From left to right: Input, GauGAN~\cite{park2019semantic}, CC-FPSE~\cite{liu2019learning}, DAGAN (Ours) and GT. These samples were randomly selected without cherry-picking for visualization purposes.}
	\label{fig:supp_ade_results1}
\end{figure*}

\begin{figure*} [!t]
	\centering
	\includegraphics[width=1\linewidth]{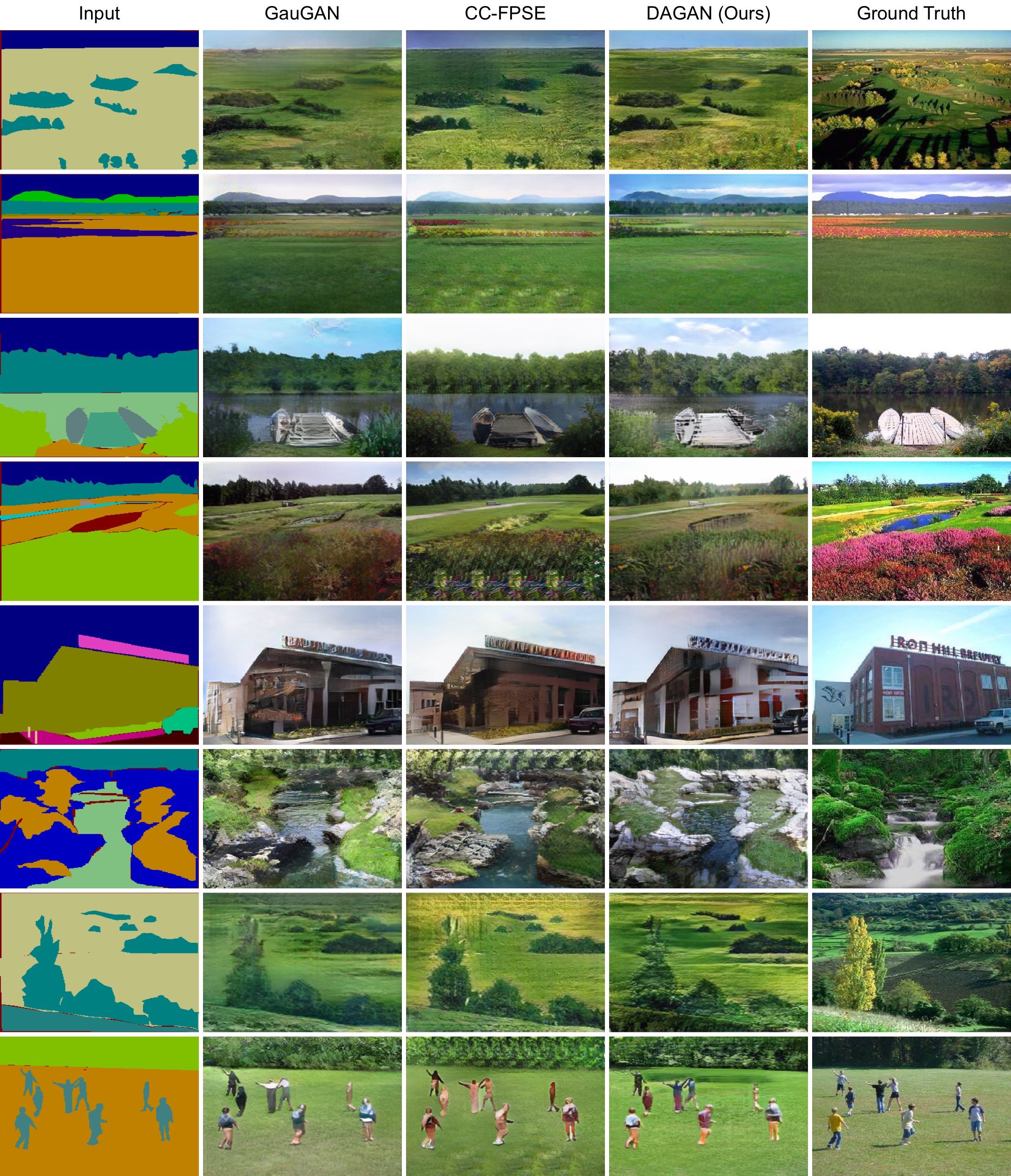}
	\caption{Qualitative comparison on ADE20K. From left to right: Input, GauGAN~\cite{park2019semantic}, CC-FPSE~\cite{liu2019learning}, DAGAN (Ours) and GT. These samples were randomly selected without cherry-picking for visualization purposes.}
	\label{fig:supp_ade_results2}
\end{figure*}

In this section, we show more generation results of the proposed DAGAN compared with those from the leading semantic image synthesis models, i.e., Pix2pixHD \cite{wang2018high}, GauGAN \cite{park2019semantic} and CC-FPSE~\cite{liu2019learning}.
The results of Cityscapes \cite{cordts2016cityscapes}, Facades \cite{tylevcek2013spatial}, CelebAMask-HQ \cite{CelebAMask-HQ}, and ADE20K \cite{zhou2017scene} are shown in Fig. \ref{fig:supp_city_results}, \ref{fig:supp_facades_results}, \ref{fig:celeba_results_supp1}, \ref{fig:celeba_results_supp2}, \ref{fig:supp_ade_results1}, and \ref{fig:supp_ade_results2}. 
We observe that the proposed DAGAN achieves visually better results than the competing methods on all the four datasets.

\section{Visualization of Learned Attention Maps}
\label{sec:s2}
\begin{figure*} [!t]
	\centering
	\includegraphics[width=1\linewidth]{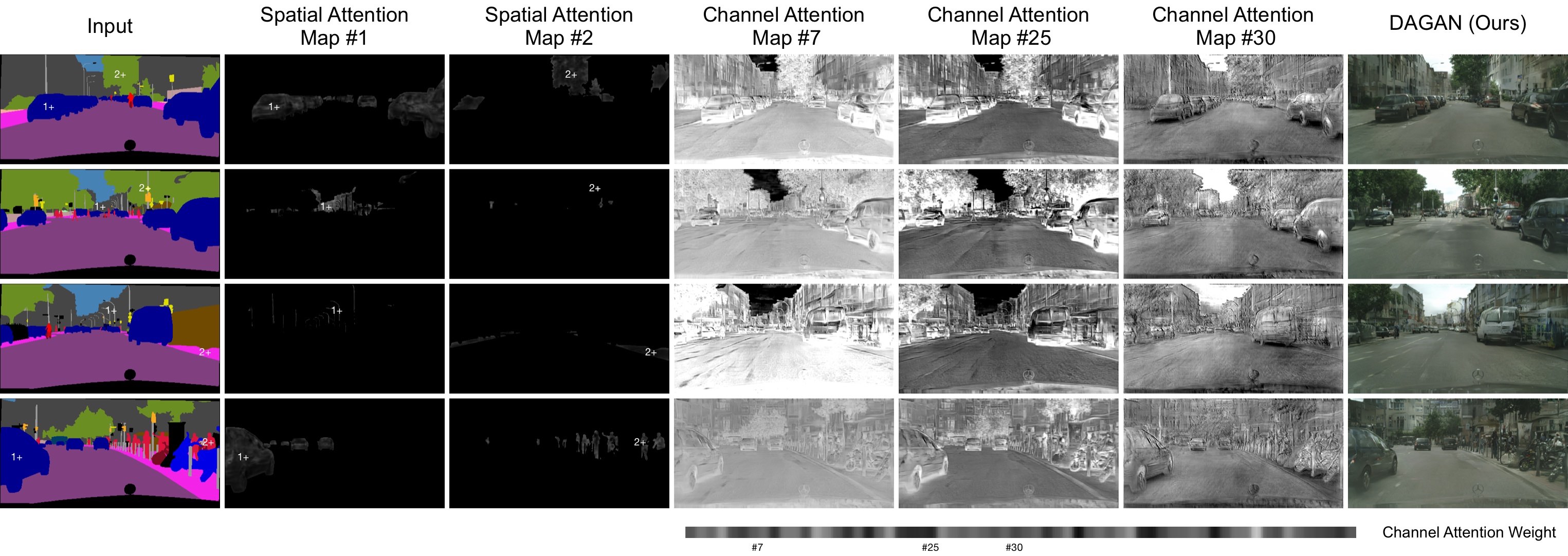}
	\caption{Visualization of learned spatial and channel attention maps on Cityscapes.
	}
	\label{fig:supp_atte}
\end{figure*}

In Fig.~\ref{fig:supp_atte} we present the learned spatial and channel attention maps. 
We observe that the spatial attention module captures global relationships within each semantic class. 
For instance, in the first row, the point `2+' is marked on a tree and its spatial attention map (in column 3) highlights most of the areas where the trees are. 
In the fourth row, the point `2+' is marked on a person and its spatial attention map (in column 3) highlights most of the areas where the people are. 

Moreover, we see that the difference of each channel is noticeable after going through the proposed channel attention module. 
For example, the $7^{th}$ channel map is brighter than both $25^{th}$ and $30^{th}$ channel maps, which means the $7^{th}$ channel map contains more information and it is more important than both $25^{th}$ and $30^{th}$ channel maps.
Both visualization results confirm the design motivation of the proposed DAGAN.

\section{Visualization of Generated Semantic Maps}
\label{sec:s3}
\begin{figure*}
	\includegraphics[width=\textwidth]{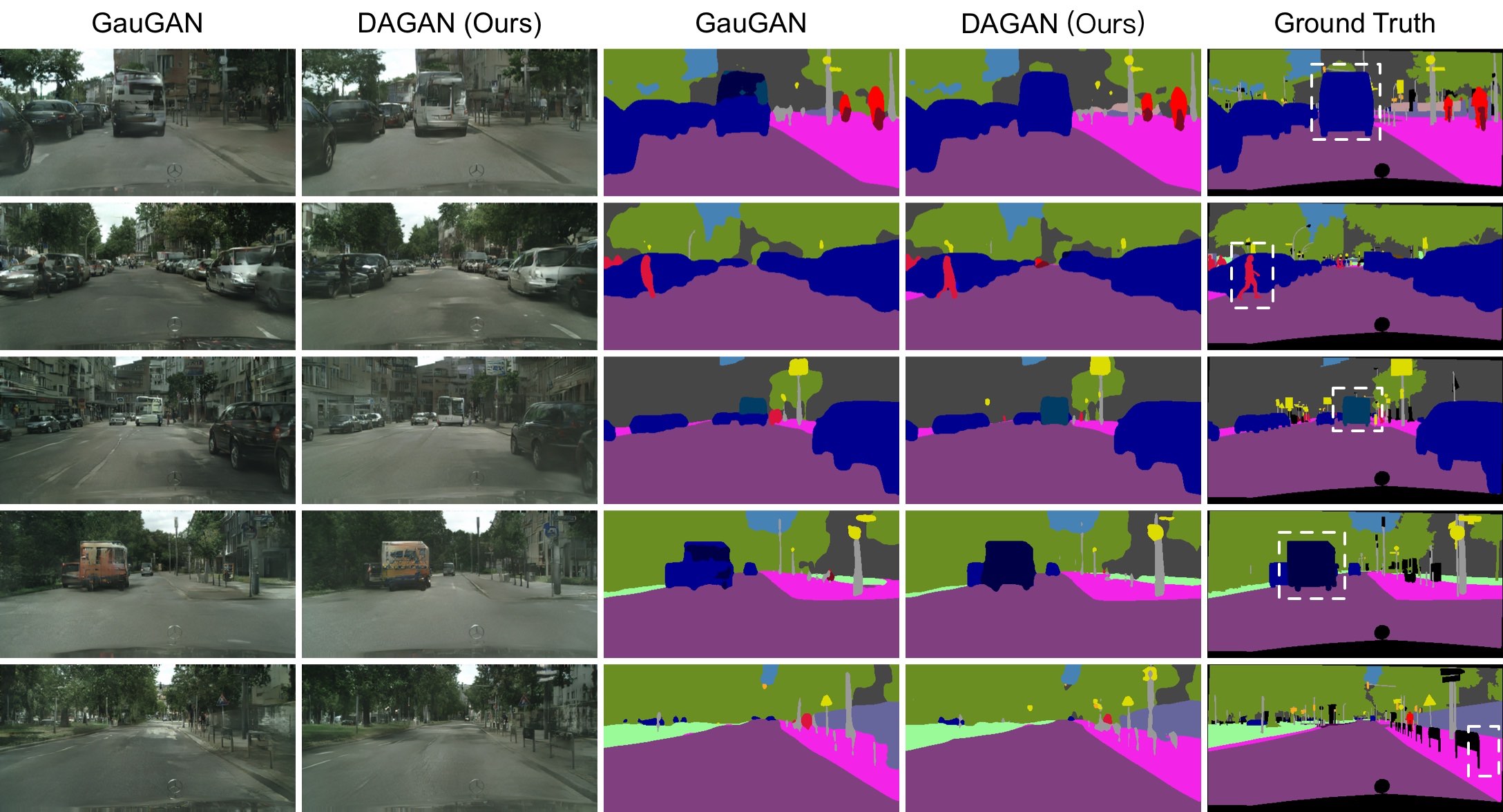}
	\caption{Visualization of generated semantic maps compared with those from GauGAN \cite{park2019semantic} on Cityscapes. These samples were randomly selected without cherry-picking for visualization purposes. Most improved regions are highlighted in the ground truths with white dash boxes.}
	\label{fig:supp_city_seg}
\end{figure*}

\begin{figure*}
	\includegraphics[width=\textwidth]{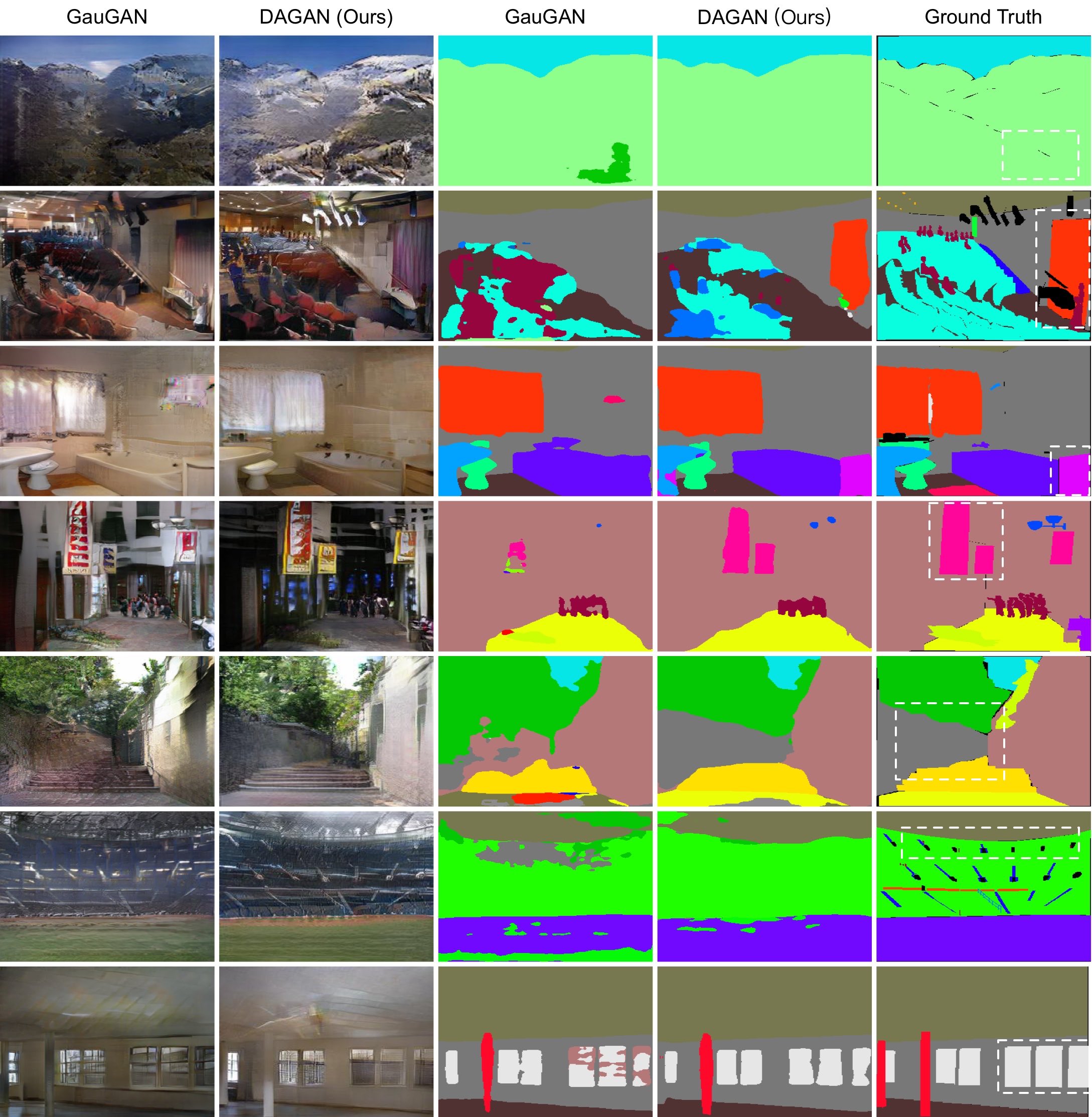}
	\caption{Visualization of generated semantic maps compared with those from GauGAN \cite{park2019semantic} on ADE20K. These samples were randomly selected without cherry-picking for visualization purposes. Most improved regions are highlighted in the ground truths with white dash boxes.}
	\label{fig:supp_ade_seg1}
\end{figure*}

\begin{figure*}
	\includegraphics[width=\textwidth]{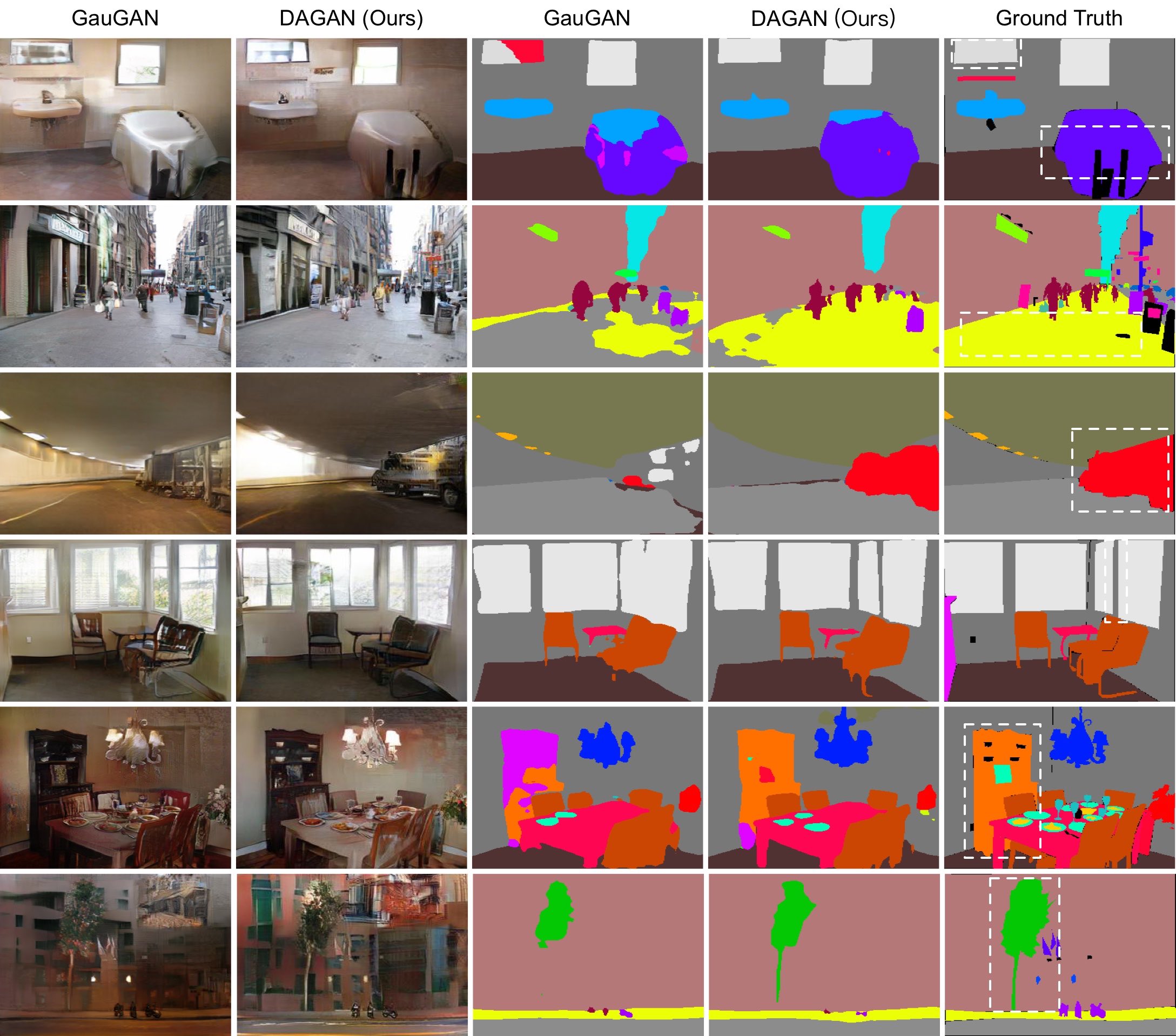}
	\caption{Visualization of generated semantic maps compared with those from GauGAN \cite{park2019semantic} on ADE20K. These samples were randomly selected without cherry-picking for visualization purposes. Most improved regions are highlighted in the ground truths with white dash boxes.}
	\label{fig:supp_ade_seg2}
\end{figure*}

We follow GauGAN \cite{park2019semantic} and use the state-of-the-art segmentation networks on the generated images to produce semantic maps: DRN-D-105 \cite{yu2017dilated} for Cityscapes and UperNet101 \cite{xiao2018unified} for ADE20K. 
The generated semantic maps of the proposed DAGAN, GauGAN, and the ground truth on Cityscapes \cite{cordts2016cityscapes} and ADE20K \cite{zhou2017scene} datasets are shown in Fig.~\ref{fig:supp_city_seg}, \ref{fig:supp_ade_seg1} and \ref{fig:supp_ade_seg2}, respectively. 
We observe that the proposed DAGAN generates more semantically-consistent results than GauGAN, further validating our motivation.

